\documentclass[a4paper]{article}
\usepackage{geometry}

\usepackage{natbib}
\usepackage{mathtools}
\mathtoolsset{showonlyrefs}  
\usepackage{amsmath,amssymb,amsthm}
\usepackage{xcolor}
\usepackage{url}
\usepackage{hyperref}
\usepackage{bm}
\usepackage{dsfont}
\usepackage{enumitem}

\newcommand*{\T}{^{\mkern-1.5mu\mathsf{T}}} %
\newcommand{\E}{\mathbb{E}}   %
\newcommand{\Q}{\mathbb{Q}}   %
\newcommand{\R}{\mathbb{R}}   %
\newcommand{\X}{\mathcal{X}}  %
\newcommand{\ip}[2]{\left\langle{#1}\right\rangle_{#2}} %
\newcommand{\norm}[2]{\left\|{#1}\right\|_{#2}} %
\newcommand{\nys}{\text{Nys}} %
\newcommand{\opnorm}[1]{\left\|{#1}\right\|_{\mathrm{op}}} %
\newcommand{\hsnorm}[2]{\left\|{#1}\right\|_{#2\otimes#2}} %
\newcommand{\tb}{\textbf}    %
\newcommand{\fm}[1]{h_p\left(\cdot,{#1}\right)} %
\newcommand{\fmc}[1]{\bar h_p\left(\cdot,{#1}\right)} %
\renewcommand{\H}{\mathcal{H}} %
\renewcommand{\O}{\mathcal{O}} %
\renewcommand{\P}{\mathbb{P}} %
\renewcommand{\b}{\mathbf}    %
\renewcommand{\d}{\mathrm{d}} %
\newcommand{\p}{\partial} %
\renewcommand{\L}{\mathcal{L}} %

\DeclareMathOperator{\trace}{tr}

\newcommand{\xis}{\left(\b x_i\right)_{i=1}^n}

\newcommand{\Span}{\mathrm{span}} %

\newtheorem{lemma}{Lemma}
\newtheorem{corollary}{Corollary}
\newtheorem{example}{Example}
\newtheorem{remark}{Remark}
\newtheorem{theorem}{Theorem}
\newtheorem{assumption}{Assumption}

\newtheorem{theoremA}{Theorem}[section] %
\newtheorem{lemmaA}[theoremA]{Lemma} %

\setenumerate{labelindent=0em,leftmargin=1.7em,topsep=0cm,partopsep=0cm,parsep=0cm,itemsep=1mm}
\setitemize{labelindent=0em,leftmargin=1.3em,topsep=0cm,partopsep=0cm,parsep=0cm,itemsep=1mm}

\title{Nyström Kernel Stein Discrepancy}

\author{%
  Florian Kalinke$^{1}$,\; Zoltán Szabó$^{2}$,\; Bharath K. Sriperumbudur$^{3}$\\
  $^{1}$Karlsruhe Institute of Technology, Karlsruhe, Germany\\
  $^{2}$London School of Economics, London, UK \\
  $^{3}$The Pennsylvania State University, University Park, PA 16802, USA\\
  \texttt{florian.kalinke@kit.edu}\hspace{0.5cm} \texttt{z.szabo@lse.ac.uk}\hspace{0.5cm} \texttt{bks18@psu.edu}
}
\date{}

\begin{document}
\maketitle

\begin{abstract}
  Kernel methods underpin many of the most successful approaches in data science
  and statistics, and they allow representing probability measures as elements
  of a reproducing kernel Hilbert space without loss of information. Recently, the kernel Stein discrepancy
  (KSD), which combines Stein's method with the flexibility of kernel techniques, gained considerable
  attention. Through the Stein operator, KSD allows the construction of
  powerful goodness-of-fit tests where it is sufficient to know the target distribution up to a multiplicative constant. However, the typical U- and V-statistic-based KSD estimators
  suffer from a quadratic runtime complexity, which hinders their application in
  large-scale settings. In this work, we propose a Nyström-based KSD acceleration---with runtime $\O\!\left(mn+m^3\right)$ for $n$ samples and $m\ll n$ Nyström points---,
  show its $\sqrt{n}$-consistency with a classical sub-Gaussian assumption, and
  demonstrate its applicability for goodness-of-fit testing on a suite of
  benchmarks. We also show the $\sqrt n$-consistency of the quadratic-time KSD estimator.
\end{abstract}

\section{INTRODUCTION}

The kernel mean embedding, which  involves mapping probability distributions into a reproducing kernel
Hilbert space (RKHS; \citealt{aronszajn50theory}) has found various far-reaching applications in the last $20$ years. For example, it allows to measure the discrepancy between probability distributions through maximum mean discrepancy (MMD; \citealt{smola07hilbert,gretton12kernel}), defined as the distance between the corresponding mean embeddings, which underpins powerful two-sample tests. MMD is also known as energy distance \citep{szekely04testing,szekely05new,baringhaus04new} in the statistics literature; see \citet{sejdinovic13equivalence} for the equivalence. We refer to \citep{muandet17kernel} for a recent overview of kernel mean embeddings.

In addition to two-sample tests, testing for goodness-of-fit (GoF; \citealt{ingster03goftesting,lehmann21testing}) is also of central importance in data science and statistics, which involves testing $H_0:\Q=\P$ vs.\ $H_1 : \Q\neq \P$ based on samples from an unknown sampling distribution $\Q$ and a (fixed known) target distribution $\P$. %
Classical GoF tests, e.g., the Kolmogorov-Smirnov test \citep{kolmogorov33sulla,smirnov48gof}, or the test for normality by \citet{barinhaus88goftest}, usually require explicit knowledge of the target distribution. However, in practical applications, the target distribution is frequently only known up to a normalizing constant. Examples include validating the output of Markov Chain Monte Carlo (MCMC) samplers \citep{welling11mcmc,bardenet14mcmc,korattikara14mcmc}, or assessing deep generative models \citep{koller09graphicalmodels,salakhutdinov15learning}. In all these examples, one desires a powerful test, even though the normalization constant might be difficult to obtain.

A recent approach to tackle GoF testing involves applying a Stein operator \citep{stein72bound,chen21stein,anastasiou23stein} to functions in an RKHS and using them as test functions to measure the discrepancy between distributions, referred to as
kernel Stein discrepancies (KSD; \citealt{chwialkowski16kernel,liu16kernelized}). An empirical estimator of KSD can be used as a test statistic to address the GoF problem. %
In particular, the Langevin Stein operator \citep{gorham15measuring,chwialkowski16kernel,liu16kernelized,oates17control,gorham17measuring} in combination with the kernel mean embedding gives rise to a KSD on the Euclidean space $\R^d$, which we consider in this work.
As a test statistic, KSD has many desirable properties. In particular, KSD requires only knowledge of the derivative of the score function of the target distribution --- implying that KSD is agnostic to the normalization of the target and therefore does not require solving, either analytically or numerically, complex normalization integrals in Bayesian settings. This property has led to its widespread use, e.g., for assessing and improving sample quality \citep{gorham15measuring,chen18stein,chen19stein,futami19bayesian,gorham20stochastic}, validating MCMC methods \citep{coullon23mcmc}, comparing deep generative models \citep{lim19kernel}, detecting out-of-distribution inputs \citep{nalisnick19outofdist}, assessing Bayesian seismic inversion \citep{izzatullah20bayesian}, modeling counterfactuals \citep{martinez23counterfactual}, and explaining predictions \citep{sarvmaili24explaining}.
GoF testing with KSDs has been explored on Euclidean data \citep{liu16kernelized,chwialkowski16kernel}, discrete data \citep{yang18discrete}, point processes \citep{yang19point}, time-to-event data \citep{fernandez20ksdgof}, graph data \citep{xu21gofgraph}, sequential models \citep{baum23ksd}, and functional data \citep{wynne24spectral}.
The KSD statistic has also been extended to the conditional case~\citep{jitkrittum20testing}.

Estimators for Langevin Stein operator-based KSD exist. But, the classical U-statistic- \citep{liu16kernelized} and V-statistic-based \citep{chwialkowski16kernel} estimators have a runtime complexity that scales quadratically with the number of samples of the sampling distribution, which limits their deployment to large-scale settings. %
 To address this bottleneck, \citet{chwialkowski16kernel} introduced a linear-time statistic that suffers from low statistical power compared to its quadratic-time counterpart. \citet{jitkrittum17lineartime} proposed the finite set Stein discrepancy (FSSD), a linear-time approach that replaces the RKHS-norm by the $L_2$-norm approximated by sampling; the sampling can either be random (FSSD-rand) or optimized w.r.t.\ a power proxy (FSSD-opt). Another approach \citep{huggins18random} is employing the random Fourier feature (RFF; \citealt{rahimi07random,sriperumbudurszabo15optimal}) method to accelerate the KSD estimation. However, it is known \citep[Proposition~1]{chwialkowski15fast} that the resulting statistic fails to distinguish a large class of measures.
\citet{huggins18random} generalize the idea of replacing the RKHS-norm by going from $L_2$-norms to $L_p$ ones, to obtain feature Stein discrepancies. They present an efficient approximation, random feature Stein discrepancies (RFSD), which is a (near-)linear time estimator. However, successful deployment of the method depends on a good choice of parameters, which, while the authors provide guidelines, can be challenging to select and tune in practice.

Our work alleviates these severe bottlenecks. We employ the Nyström method \citep{williams01using} to accelerate KSD estimation and show the $\sqrt n$-consistency of our proposed estimator. The main technical challenge is that the Stein kernel (induced by the Langevin Stein operator and the original kernel) is typically unbounded while existing statistical Nyström analysis \citep{rudi15less,chatalic22nystrom,sterge22nystroem,kalinke23nystrom,chatalic25nystroem} usually considers bounded kernels. To tackle unbounded kernels, we select a classical sub-Gaussian assumption, which we impose on the feature map associated to the kernel, and show that existing methods of analysis can successfully be extended to handle this novel case.
In this sense, our work, besides  \citet{della21regularized}, which requires a similar sub-Gaussian condition for analyzing empirical risk minimization on random subspaces, is a first step in analyzing the consistency of the unbounded case in the Nyström setting.

Our main \tb{contributions} are the following.
\begin{enumerate}
  \item We introduce a Nyström-based acceleration of kernel Stein discrepancy. The proposed estimator runs in $\O\left(mn + m^3\right)$ time, with $n$ samples and $m \ll n$ Nyström points.
  \item We prove the $\sqrt n$-consistency of our estimator in a classical sub-Gaussian setting, which extends (in a non-trivial fashion) existing results for Nyström-based methods \citep{rudi15less,chatalic22nystrom,sterge22nystroem,kalinke23nystrom} focusing on bounded kernels.
  \item We perform an extensive suite of experiments to demonstrate the applicability of the proposed method.
  Our proposed approach achieves competitive results throughout all experiments.
\end{enumerate}

The paper is structured as follows. We introduce the notations used throughout the article (Section~\ref{sec:notations}) followed by recalling the classical quadratic-time KSD estimators (Section~\ref{sec:problem-formulation}). In Section~\ref{sec:nys-estimator}, we detail our proposed Nyström-based estimator, alongside with its adaptation to a modified wild bootstrap goodness-of-fit test (Section~\ref{sec:nys-testing}), and our theoretical guarantees (Section~\ref{sec:nys-guarantees}).  Experiments demonstrating the efficiency of our  Nyström-KSD estimator are provided in Section~\ref{sec:experiments}. Limitations are in Section~\ref{sec:limitations}. Proofs and additional experiments are deferred to the appendices.

\section{NOTATIONS}
\label{sec:notations}
In this section, we introduce our notations 
$[N]$,
$\lesssim$, $\gtrsim$, $\asymp$,
$\mathds 1_{A}$,
$\{\{\cdot\}\}$,
$\bm 1_n$, $\bm 0_n$
$\b I_n$,
$\b A^-$,
$\b A\T$,
$\b A^{-1}$,
$\nabla_{\b x}$,
$\mathcal M_1^+(\X)$,
$\P^n$,
$\P_1\otimes \P_2$,
$\O_\P(\cdot)$,
$B(\H_k)$,
$k$,
$\H_k$,
$P_U$,
$\opnorm{\cdot}$,
$\mathcal L(\H_k)$,
$A^*$,
$A^{\frac{1}{2}}$,
$\trace$,
$\mu_k$,
$f\otimes g$,
$C_{\P,k}$,
$C_{\P,k,\lambda}$,
$\mathcal N_{\P,k}$,
$\norm{\cdot}{L_r(\P)}$,
$\norm{\cdot}{\psi_r}$.

Let $[N] := \{1,\ldots,N\}$ for a positive integer $N$.
For $a_1,a_2 \ge 0$, $a_1 \lesssim a_2$ (resp.\ $a_1 \gtrsim a_2$) means that $a_1\le ca_2$ (resp.\ $a_1\ge c'a_2$) for an absolute constant $c>0$ (resp.\ $c' > 0$), and we write $a_1 \asymp a_2$ iff.\ $a_1\lesssim a_2$ and $a_1\gtrsim a_2$. We write $\mathds 1_{A}$ for the indicator function of a set $A$ and $\left\{\left\{\cdot\right\}\right\}$ for a multiset.
The $n$-dimensional vector of ones is denoted by
$\bm 1_n = \left(1,\ldots,1\right)\T \in\R^n$ , that of $n$ zeros by $\bm 0_n = \left(0,\ldots,0\right)\T \in \R^n$. The identity matrix is  $\b I_n\in\R^{n\times n}$. For a matrix $\b A \in\R^{d_1\times d_2}$, $\b A^- \in \R^{d_2\times d_1}$ denotes its (Moore-Penrose) pseudo-inverse, and $\b A\T \in \R^{d_2\times d_1}$ stands for the transpose of $\b A$. We write $\b A^{-1}\in \R^{d\times d}$ for the inverse of a non-singular matrix $\b A \in \R^{d\times d}$. For a differentiable function $f:\R^d \rightarrow \R$ , let $\nabla_{\b x}f(\b x)=\left(\frac{\p f(\b x)}{\p x_i}\right)_{i=1}^d \in \R^d$.

Let $\left(\X,\tau_\X\right)$ be a topological space and $\mathcal B\left(\tau_\X\right)$ the corresponding Borel $\sigma$-algebra. Probability measures considered in this article are meant w.r.t.\ the measurable space $\left(\X, \mathcal B\left(\tau_\X\right)\right)$ and are written as $\mathcal M_1^+\left(\X\right)$; for instance, the set of Borel probability measures on $\R^d$ is $\mathcal M_{1}^{+}\left(\R^d\right)$. The $n$-fold product measure of $\P\in \mathcal M_1^+\left(\X\right)$ is denoted by $\P^n \in \mathcal M_1^+\left(\X^n\right)$. The product of $\P_1 \in \mathcal M_1^+(\X_1)$ and $\P_2 \in \mathcal M_1^+(\X_2)$ is written as $\P_1\otimes \P_2$  ($\in \mathcal M_1^+(\X_1\times \X_2)$), where $\left(\X_1,\tau_{\X_1}\right)$ and $\left(\X_2,\tau_{\X_2}\right)$ are topological spaces.
For a sequence of i.i.d.\ real-valued random variables $X_n \sim \P \in \mathcal M_1^+(\R)$ and a sequence of positive $r_n$-s, $X_n = \O_\P(r_n)$ means that $\frac{X_n}{r_n}$ is bounded in probability.
The unit ball in a Hilbert space $\H$ is denoted by $B(\H)=\left\{f\in \H \mid \left\|f\right\|_{\H}\le 1 \right\}$. The reproducing kernel Hilbert space with $k:\R^d\times\R^d\rightarrow\mathbb{R}$ as the reproducing kernel is denoted by $\H_k$.
Throughout the paper, $k$ is assumed to be measurable and  $\H_{k}$ to be separable.\footnote{For instance, a continuous kernel $k:\R^d \times \R^d \to \R$ implies both properties; see \citet[Lemma~4.33]{steinwart08support} for separability.} %
Given a closed linear subspace $U\subseteq \H_k$, the (orthogonal) projection of $h\in\H_k$ on $U$ is denoted by $P_Uh\in U$;
$u=P_Uh$ is the unique vector such that $h-u \perp U$.
For any $u\in U$, $\norm{h-P_{U}h}{\H_k}\le \norm{h-u}{\H_k}$, that is, $P_Uh$ is the closest element in $U$ to $h$. A linear operator $A : \H_k \to \H_k$ is called bounded if  $\opnorm{A} := \sup_{\norm{h}{\H_k}=1}\norm{Ah}{\H_k} < \infty$; the set of $\H_k\rightarrow \H_k$ bounded linear operators is denoted by $\L(\H_k)$. An $A \in \L(\H_k)$ is called positive (shortly $A\ge 0$) if  it is self-adjoint ($A^*=A$, where $A^*\in \L(\H_k)$ is defined by $\langle Af,g \rangle_{\H_k} = \langle f,A^*g \rangle_{\H_k}$ for all $f,g\in \H_k$), and $\langle Ah,h\rangle_{\H_k} \ge 0$ for all $h\in \H_k$. If $A\ge 0$, then there exists a unique $B \ge 0$ such that $B^2 = A$; we write $B = A^{\frac{1}{2}}$ and call $B$ the square root of $A$. An $A \in \L(\H_k)$ is called trace-class if $\sum_{i\in I} \langle (A^*A)^{\frac{1}{2}}e_i,e_i \rangle_{\H_k} <\infty$ for some countable orthonormal basis (ONB) $(e_i)_{i\in I}$ of $\H_k$, and in this case $\trace(A):=\sum_{i\in I} \langle Ae_i,e_i \rangle_{\H_k} <\infty$.\footnote{The trace-class property and the value of $\trace(A)$ is independent of the specific ONB chosen. The separability of $\H_k$ implies the existence of a countable ONB in it.} For a self-adjoint trace-class operator $A$ with eigenvalues $(\lambda_i)_{i\in I}$, $\trace(A)=\sum_{i\in I} \lambda_i$. An operator $A \in \L(\H_k)$ is called compact if $\overline{\{Ah\,|\, h\in B(\H_k)\}}$ is compact, where $\overline{\cdot}$ denotes the closure. A trace class operator is compact, and a compact positive operator $A$ has largest eigenvalue $\opnorm{A}$. For any $A\in \L(\H_k)$, it holds that  $\opnorm{A^*A} = \opnorm{A}^2$ (which is called the $C^*$ property).

The mean embedding of a probability measure $\P\in\mathcal M_1^+(\R^d)$ into the RKHS associated to kernel $k:\R^d \times \R^d \rightarrow \R$ is
  $\mu_k(\P) = \int_{\R^d} k\left( \cdot, \b x \right)\d\P(\b x) \in \H_{k}$,
where the integral is meant in Bochner's sense \citep[Chapter~II.2]{diestel77vector}. The mean element $\mu_k(\P)$ exists iff.\ $\int_{\R^d}\norm{k\left( \cdot, \b x \right)}{\H_{k}}\d\P(\b x) < \infty$ \citep[p.~45; Theorem~2]{diestel77vector}.

Let $f,g \in \H_k$. Their tensor product is written as $f\otimes g \in \H_k\otimes \H_k$, where $\H_k\otimes \H_k$ is the tensor product Hilbert space;
further, $f\otimes g : \H_k\to\H_k$ defines a rank-one operator by $h\mapsto f\ip{g,h}{\H_k}$. It is known that $\H_k\otimes\H_k$ is also an RKHS \citep[Theorem~13]{berlinet04reproducing}.
Given a probability measure $\P\in\mathcal M_1^+\left(\R^d\right)$ and a kernel $k:\R^d \times \R^d\rightarrow \R$,  the uncentered covariance operator
\begin{align}
  C_{\P,k} = \int_{\R^d}k\left( \cdot, \b x \right)\otimes k\left( \cdot, \b x \right) \d\P(\b x) \in \H_k\otimes \H_k \label{eq:covariance-operator}
\end{align}
exists if $\int_{\X}\norm{k\left( \cdot,\b x \right)}{\H_{k}}^2\d\P(\b x) < \infty$; $C_{\P,k}$ is a positive trace-class operator. We define $C_{\P,k,\lambda} = C_{\P,k}+\lambda I$, where $I$ denotes the identity operator and $\lambda >0$.
The effective dimension of $\P\in\mathcal M_1^+\left(\R^d\right)$ is defined as $\mathcal N_{\P,k}(\lambda) :%
= \trace \left( C_{\P,k}C_{\P,k,\lambda}^{-1} \right)\le\frac{\trace\left(C_{\P,k}\right)}{\lambda}$.\footnote{\label{fn:cov-ineq} This inequality is implied by $\trace\left(C_{\P,k}C_{\P,k,\lambda}^{-1}\right) = \sum_{i  \in I} \frac{\lambda_i}{\lambda_i+\lambda}\le  \frac{1}{\lambda}\sum_{i\in I}\lambda_i = \frac{\trace\left(C_{\P,k}\right)}{\lambda}$, where $(\lambda_i)_{i\in I}$ denote the eigenvalues of $C_{\P,k}$.}
With $r\ge 1$ and a real-valued random variable $X: \left(\Omega,\mathcal{A},\P\right) \rightarrow \left(\R, \mathcal{B}(\tau_\R)\right)$, where $\mathcal{B}(\tau_\R)$ denotes the Borel $\sigma$-field on $\R$, let $\left\|X\right\|_{L_r(\P)}= \left[\int_{\Omega} |X(\omega)|^r \d \P (\omega)\right]^{\frac{1}{r}}$. For $r \in \{1,2\}$, let $\psi_r(u) = e^{u^r}-1$ and $\norm{X}{\psi_r} := \inf\left\{C > 0 \mid \E_{X\sim\P}\psi_r\left(\frac{|X|}{C}\right) \le 1\right\}$.
A real-valued random variable $X\sim\P\in\mathcal M_1^+(\R)$ is called sub-exponential if $\norm{X}{\psi_1}<\infty$ and sub-Gaussian if $\norm{X}{\psi_2} < \infty$.
In the following, we specialize Definition~2 by \citet{koltchinskii17concentration} stated for Banach spaces to (reproducing kernel) Hilbert spaces by using the Riesz representation theorem.
A centered $\H_k$-valued random variable $X\sim\Q\in\mathcal M_1^+\left( \H_k \right)$ is called sub-Gaussian iff.\ there exists a universal constant $C>0$ such that for all $u \in \H_k$:
\begin{align}
  \norm{\ip{X,u}{\H_k}}{\psi_2} \le C \norm{\ip{X,u}{\H_k}}{L_2(\Q)} <\infty. \label{eq:sub-gaussian}
\end{align}

\section{PROBLEM FORMULATION}
\label{sec:problem-formulation}

We now introduce our quantity of interest, the kernel Stein discrepancy. Let $\H_{k}^d:=\times_{i=1}^d\H_k$ be the product RKHS  with inner product defined by $\ip{\b f,\b g}{\H_k^d} = \sum_{i=1}^d\ip{f_i,g_i}{\H_k}$ for $\b f=\left( f_i \right)_{i=1}^d, \b g = \left( g_i \right)_{i=1}^d \in \H_k^d$.
Let $\P, \Q \in\mathcal M_1^+\left(\R^{d}\right)$ be fixed; we refer to $\P$ as the target distribution and to $\Q$ as the sampling distribution. Assume that $\P$ is absolutely continuous w.r.t.\ the Lebesgue measure and let $p$ be the corresponding density (w.r.t.\ Lebesgue measure).
We assume that $p$ is continuously differentiable with support $\R^d$, $p(\b x) >0$ for all $\b x \in \R^d$, and $\lim_{\norm{\b x}{}\to\infty}f(\b x)p(\b x) = 0$ for all $f \in \H_k$. The last property holds for instance if $p$ is bounded and $\lim_{\norm{\b x}{}\to\infty}f(\b x) = 0$ for all $f \in \H_k$. Further, we assume that $k$ is continuously differentiable in both arguments. This condition will imply the  measurability of $h_p$ and the separability of $\H_{h_p}$, both quantities defined below.
The Stein operator \citep[(4)]{gorham15measuring} is defined as
$\left(T_p \b f\right)(\b x) = \langle \nabla_{\b x} [\log p(\b x)], \b f(\b x)\rangle + \sum_{i=1}^d \dfrac{\partial f_i(\b x)}{\partial x_i}$ $\left(\b f \in \H_k^d,\,\b x \in \R^d\right)$.
With this definition at hand,
\begin{align}
    \left(T_p \b f\right)(\b x) &= \ip{\b f, \bm \xi_p(\b x)}{\H_k^d},\\
\bm \xi_p(\b x) &= \left[\nabla_{\b x}\left( \log p(\b x)\right)k\left( \cdot, \b x \right)  + \nabla_{\b x} k\left( \cdot, \b x \right) \right] \in \H_{k}^{d}  \label{eq:xi-p}
\end{align}
for all $\b f \in \H_k^d$ and $\b x \in \R^d$, with kernel (for $\b x, \b y \in \R^d$)
\begin{align}
  h_p(\b x, \b y) &= \ip{\bm \xi_p(\b x),\bm \xi_p(\b y)}{\H_{k}^{d}}\\
  &= \langle h_p(\cdot,\b x),h_p(\cdot, \b y)\rangle_{\H_{h_p}}; \label{eq:h-p}
\end{align}
notice that $\bm \xi_p(\b x)$ and $h_p(\cdot,\b x)$ map to different feature spaces ($\H_k^d$ and $\H_{h_p}$, respectively) but yield the same kernel $h_p$, which, with \eqref{eq:xi-p}, takes the explicit form
\begin{align}
   h_p(\b x, \b y) &= \ip{\nabla_{\b x}\log p(\b x),\nabla_{\b y}\log p(\b y)}{\R^d}k(\b x, \b y) 
  + \ip{\nabla_{\b y}\log p(\b y),\nabla_{\b x}k(\b x, \b y)}{\R^d} \\
  &\quad + \ip{\nabla_{\b x}\log p(\b x),\nabla_{\b y}k(\b x, \b y)}{\R^d} + \sum_{i=1}^d\dfrac{\partial^2k(\b x,\b y)}{\partial x_i\partial y_i}. \notag
\end{align}

The kernel Stein discrepancy (KSD; \citealt{chwialkowski16kernel,liu16kernelized}) then is defined as an integral probability metric \citep{zolotarev83probability,muller97integral}
\begin{align}
  S_{p}(\Q) &= \sup_{\b f \in B\left(\H_k^d\right)}\underbrace{\E_{X\sim\P}\left[T_p\b f(X)\right]}_{\stackrel{(a)}{=}0} -\; \E_{X\sim\Q}\left[T_p\b f(X)\right]
 =\sup_{\b f \in B\left(\H_k^d\right)}\ip{\b f,\E_{X\sim\Q}\bm \xi_p(X)}{\H_k^d} \\
 &= \norm{\E_{X\sim \Q}\bm \xi_p(X)}{\H_k^d} \stackrel{(b)}{=}\norm{\E_{X\sim\Q}\fm{X}}{\H_{h_p}}, \;\;
 \label{eq:stein-discrepancy}
\end{align}
where (a) holds by the construction of KSD and (b) follows from \eqref{eq:h-p}.

Given a sample $\hat \Q_n = \{\b x_i\}_{i=1}^n \sim \Q^n$, the popular V-statistic-based estimator \citep[Section~2.2]{chwialkowski16kernel}  is obtained by replacing $\Q$ with the empirical measure $\hat \Q_n$; it takes the form
\begin{align}
  S^2_p\left(\hat \Q_n\right) = \frac{1}{n^2}\sum_{i,j=1}^nh_p(\b x_i,\b x_j), \label{eq:ksd-quad-time-v}
\end{align}
and can be computed in $\O\!\left(n^2\right)$ time.
The corresponding U-statistic-based estimator \citep[(14)]{liu16kernelized} has a similar expression but omits the diagonal terms, that is, 
\begin{align}
  S^2_{p,u}\left(\hat \Q_n\right) = \frac{1}{n(n-1)}\sum_{1\le i\neq j \le n}^nh_p(\b x_i,\b x_j);
\end{align}
it also has a runtime cost of $\O\!\left(n^2\right)$.
For large-scale applications, the quadratic runtime is a significant bottleneck; this is the shortcoming we tackle in the following.

\section{PROPOSED NYSTRÖM-KSD}
\label{sec:prop-ksd-estim}
To enable the efficient estimation of \eqref{eq:stein-discrepancy}, we propose a Nyström technique-based estimator in Section~\ref{sec:nys-estimator} and an accelerated wild bootstrap test in Section~\ref{sec:nys-testing}. In Section~\ref{sec:nys-guarantees}, our consistency results are collected.

\subsection{The Nyström-KSD Estimator}\label{sec:nys-estimator}

We consider a subsample $\tilde \Q_m = \left\{\left\{\tilde {\b x}_1,\ldots,\tilde {\b x }_m\right\}\right\}$ of size $m$ (sampled with replacement), the so-called Nyström sample, of the original sample $\hat \Q_n = \left\{\b x_1,\ldots,\b x_n\right\}$; the tilde indicates a relabeling. 
The best approximation of $S_p(\Q)$ in RKHS-norm-sense, when using $m$ Nyström samples, can be obtained by considering the orthogonal projection of $\E_{X\sim\Q}\fm X$ onto
$\H_{h_p,m}  :=\Span\left\{\fm{\tilde {\b x}_i}\mid i \in [m]\right\} \subset \H_{h_p}$, with feature map $\fm{\tilde {\b x}_i}$ and associated kernel $h_{p}$  defined in \eqref{eq:h-p}.
As $\Q$ is unknown in practice and only available via samples $\hat\Q_n \sim\Q^n$, we consider the orthogonal projection of $\E_{X\sim\hat \Q_n}\fm X$ onto $\H_{h_p,m}$ instead. In other words, we aim to find the weights $\bm \alpha = (\alpha_i)_{i=1}^m \in \R^m$ that correspond to the minimum norm solution of the cost function
\begin{align}
  \min_{\bm\alpha \in \R^m}\Bigg\|\underbrace{\frac1n\sum_{i=1}^n\fm{\b x_i}}_{=\E_{X\sim\hat\Q_n}\fm X}-\sum_{i=1}^m\alpha_i\fm{\tilde {\b x}_i}\Bigg\|_{\H_{h_p}}, \label{eq:optim-projection}
\end{align}
which gives rise to the squared KSD estimator\footnote{$\tilde S_p^2\left(\hat\Q_n\right)$ indicates dependence on $\hat \Q_n$.}
\begin{align}
\tilde S_p^2\left(\hat \Q_n\right)&:=\norm{\sum_{i=1}^m\alpha_i\fm{\tilde {\b x}_i}}{\H_{h_p},m}^2 
=\norm{P_{\H_{h_p,m}}\E_{X\sim\hat\Q_n}\fm X}{\H_{h_p,m}}^2.
\label{eq:KSD-estimator}
\end{align}

\begin{lemma}[Nyström-KSD Estimator] \label{lemma:nystroem-ksd-estimator}
  The squared KSD estimator \eqref{eq:KSD-estimator} takes the form
  \begin{align}
     \tilde S_p^2\left(\hat \Q_n\right) &= \bm\beta_{p}\T\b K_{h_p,m,m}^{-}\bm\beta_{p},\label{eq:nystroem-ksd-estimator}
  \end{align}
  where   $\bm\beta_p = \frac1n\b K_{h_p,m,n} \bm1_{n}\in \R^m$, Gram matrix $\b K_{h_p,m,m} = \left[h_p\left(\tilde {\b x}_i,\tilde {\b x}_j\right)\right]_{i,j=1}^m \in \R^{m\times m}$, and $\b K_{h_p,m,n} = \left[h_p\left(\tilde {\b x}_i,{\b x}_j\right)\right]_{i,j=1}^{m,n} \in \R^{m\times n}$.
\end{lemma}

\begin{remark}\label{remark:nystroem-ksd-lemma}~
  \begin{enumerate}[label=(\alph*)]
  \item \tb{Runtime complexity.}\label{item:speedup} The computation of \eqref{eq:nystroem-ksd-estimator} consists of calculating $\bm \beta_p$, pseudo-inverting $\bm K_{h_p,m,m}$, and obtaining the quadratic form $\bm\beta_{p}\T\b K_{h_p,m,m}^{-}\bm\beta_{p}$. The calculation of $\bm \beta_p$ requires $\O(mn)$ operations, due to the multiplication of an $m\times n$ matrix with a vector of length $n$. Inverting the $m\times m$ matrix $\bm K_{h_p,m,m}$ costs $\O(m^3)$,\footnote{Although faster algorithms for (pseudo) matrix inversion exist, we consider the runtime that one typically encounters in practice.} dominating the cost of $\O(m^2)$ needed for the computation of $\bm K_{h_p,m,m}$. The quadratic form $\bm\beta_{p}\T\b K_{h_p,m,m}^{-}\bm\beta_{p}$ has a computational cost of $\O\left(m^2\right)$. Hence, \eqref{eq:nystroem-ksd-estimator} can be computed in $\O\left(mn + m^3\right)$, which means that for $m=o\left(n^{2/3}\right)$, our proposed Nyström-KSD estimator guarantees an asymptotic speedup. %

  \item \tb{Comparison of \eqref{eq:ksd-quad-time-v} and \eqref{eq:nystroem-ksd-estimator}.} The Nyström estimator \eqref{eq:nystroem-ksd-estimator} recovers the V-statistic-based estimator \eqref{eq:ksd-quad-time-v} when no subsampling is performed and provided that $\b K_{h_p,n,n}$ is invertible.
  \item \tb{Comparison to \citet{chatalic22nystrom}.} We note that the estimator \eqref{eq:nystroem-ksd-estimator} corresponds precisely to \citet[(5)]{chatalic22nystrom}. We consider the analysis of this known estimator in the case of unbounded feature maps---which arise in the KSD setting---as one of our core contributions, which we detail in Section~\ref{sec:nys-guarantees}.
  \end{enumerate}
\end{remark}

\subsection{Nyström Bootstrap Testing}\label{sec:nys-testing}
In this section, we discuss how one can use \eqref{eq:nystroem-ksd-estimator} for accelerated goodness-of-fit testing. We recall that the goal of goodness-of-fit testing is to test $H_0 : \Q = \P$ versus $H_1 :  \Q \neq \P$, given samples $\hat \Q_n = \left\{\b x_1,\ldots,\b x_n\right\}$ and target distribution $\P$. Recall that KSD relies on score functions ($\nabla_{\b x}[\log p(\b x)]$); hence knowing $\P$ up to a multiplicative constant is enough. To use the Nyström-based estimator \eqref{eq:nystroem-ksd-estimator} for goodness-of-fit testing, we propose to obtain its null distribution by a Nyström-based bootstrap procedure. Our method builds on the existing test for the V-statistic-based KSD, detailed in~\citet[Section~2.2]{chwialkowski16kernel}, which we quote in the following. Define the bootstrapped statistic by
\begin{align}
  B_n = \frac{1}{n^2}\sum_{i,j=1}^nw_{i}w_{j}h_p\left(\b x_i,\b x_j\right), \label{eq:kacper-bootstrap}
\end{align}
with $w_i \in \{-1,1\}$ an auxiliary Markov chain defined by
\begin{align}
  w_i = \mathds 1_{(U_i > 0.5)}w_{i-1}-\mathds 1_{(U_i \le 0.5)}w_{i-1}, \label{eq:markov-chain}
\end{align}
where $U_i \stackrel{\text{i.i.d.}}{\sim} \operatorname{Unif}(0,1)$,
that is, $w_i$ changes sign with probability $0.5$. The test procedure is as follows.
\begin{enumerate}
\item Calculate the test statistic \eqref{eq:ksd-quad-time-v}.
\item Obtain $D$ wild bootstrap samples $\{B_{n,i}\}_{i=1}^D$ with \eqref{eq:kacper-bootstrap} and estimate the $1-\alpha$ empirical quantile of these samples.
\item Reject the null hypothesis if the test statistic \eqref{eq:ksd-quad-time-v} exceeds the quantile.
\end{enumerate}

\eqref{eq:kacper-bootstrap} requires $\O\!\left( n^2 \right)$ computations, which yields a total cost of $\O\! \left( Dn^2 \right)$ for obtaining $D$ bootstrap samples, rendering large-scale goodness-of-fit tests unpractical.

We propose the Nyström-based bootstrap
\begin{align}
  B_{n}^{\nys} = \frac{1}{n^2}\b w\T \b K_{h_p,n,m}\b K_{h_p,m,m}^-\b K_{h_p,m,n}\b w, \label{eq:nys-bootstrap}
\end{align}
with $\b w = \left(w_i  \right)_{i=1}^n\in\R^n$ collecting the $w_i$-s ($i\in[n]$) defined in \eqref{eq:markov-chain}; $\b K_{h_p,n,m}$ $(= \b K_{h_p,m,n}\T)$ and $\b K_{h_p,m,m}$ are defined as in Lemma~\ref{lemma:nystroem-ksd-estimator}. The approximation is based on the fact~\citep{williams01using} that $\b K_{h_p,n,m}\b K_{h_p,m,m}^-\b K_{h_p,m,n}$ is a low-rank approximation of $\b K_{h_p,n,n}$, that is, $\b K_{h_p,n,m}\b K_{h_p,m,m}^-\b K_{h_p,m,n} \approx \b K_{h_p,n,n}$.
Hence, our proposed procedure \eqref{eq:nys-bootstrap} approximates \eqref{eq:kacper-bootstrap} but reduces the cost from $\O \left( n^2 \right)$ to $\O \left(nm + m^3\right)$ if one computes from left to right (also refer to Remark~\ref{remark:nystroem-ksd-lemma}\ref{item:speedup}); in the case of $m=o \left( n^{2/3} \right)$ this guarantees an asymptotic speedup. We obtain a total cost of $\O \left( D \left( nm+m^3 \right) \right)$ for obtaining the wild bootstrap samples. This acceleration allows KSD-based goodness-of-fit tests to be applied on large data sets.

\subsection{Guarantees}
\label{sec:nys-guarantees}
This section is dedicated to the statistical behavior of the proposed Nyström-KSD estimator \eqref{eq:nystroem-ksd-estimator}.

The existing analysis of Nyström estimators \citep{rudi15less,chatalic22nystrom,sterge22nystroem,kalinke23nystrom} considers bounded kernels only. Indeed, if $\sup_{\b x\in\R^d}\norm{\fm{\b x}}{\H_{h_p}} <\infty$, the consistency of \eqref{eq:nystroem-ksd-estimator} is implied by \citet[Theorem~4.1]{chatalic22nystrom}, which we include here for convenience of comparison. In the following, we denote the randomness in the choice of Nyström samples by $(i_j)_{j=1}^m \stackrel{\text{i.i.d.}}{\sim} \text{Unif}([n]) =: \Lambda$, which means that $\tilde {\b x}_j = \b x_{i_j}$ with $j \in [m]$.

\begin{theorem}[Bounded case]
\label{thm:chatalic} Assume the setting of Lemma~\ref{lemma:nystroem-ksd-estimator}, $C_{\Q,h_p}\ne 0$, $m\ge 4$ Nyström samples, and a bounded Stein feature map ($\sup_{\b x\in\R^d}\norm{\fm{\b x}}{\H_{h_p}} =: K <\infty$). Then, for any $\delta\in(0,1)$, it holds with $\left(\Q^n\otimes \Lambda^m\right)$-probability of at least $1-\delta$ that
  \begin{align}
    \MoveEqLeft\left|S_p(\Q) - \tilde S_p\left(\hat \Q_n\right)\right| \le \frac{c_1}{\sqrt n} + \frac{c_2}{m} 
    + \frac{c_3\sqrt{\log\frac{m}{\delta}}}{m}\sqrt{\mathcal N_{\Q,h_p}\!\left(\frac{12K^2\log\frac{m}{\delta}}{m}\right)},
  \end{align}
  when $m \ge \max\left(67,12K^2\opnorm{C_{\Q,h_p}}^{-1}\right)\log(m/\delta)$, where $c_1$, $c_2$, and $c_3$ are positive constants.
\end{theorem}

However, in practice, the feature map of KSD is typically unbounded and Theorem~\ref{thm:chatalic} is not applicable, as it is illustrated in the following example with the frequently-used Gaussian kernel.
\begin{example}[KSD yields unbounded kernel]\label{example:unbounded-kernel}
  Consider univariate data ($d=1$), unnormalized target density $p(x) = e^{-x^2/2}$ (corresponding to $\P = \mathcal N(0,1)$), and (i) the RBF kernel $k(x,y) = \exp\left(-\gamma(x-y)^2\right)$ with $\gamma>0$, or (ii) the IMQ kernel $k(x,y) = \left(c^2+(x-y)^2\right)^{{-\beta}}$ with $\beta,c>0$. By using \eqref{eq:h-p}, direct calculation yields
      (i) $\norm{\xi_p(\cdot, x)}{\H_k}^2 =  x^2 + 2\gamma \stackrel{x\to\infty}{\to} \infty$ in the first, and
      (ii) $\norm{\xi_p(\cdot, x)}{\H_k}^2  = x^2 c^{2\beta}-2 \beta  c^{2(\beta-1)} \stackrel{x\to\infty}{\to} \infty$ in the second case.
\end{example}

\begin{remark}
    In fact, a more general result holds: If one considers a bounded continuously differentiable translation-invariant kernel $k$, the induced Stein kernel is only bounded provided that the target density $p(\b x)$ has tails that are no thinner than $e^{-\sum_{i=1}^d|x_i|}$ \citep[Remark~2]{hagrass24stein}, which clearly rules out Gaussian targets.
\end{remark}

For analyzing the setting of unbounded feature maps, we make the following assumption.

\begin{assumption}\label{ass:sub-gaussian} The centered Stein feature map $\fmc X = \fm X - \E_{X\sim \Q} \fm X$ with the sampling distribution $\Q \in \mathcal M_1^+\left(\R^d\right)$ is sub-Gaussian in the sense of \eqref{eq:sub-gaussian}, that is,
  \begin{align}
    \norm{\ip{\fmc X,u}{\H_{h_p}}}{\psi_2} \hspace{-0.2cm}\lesssim \norm{\ip{\fmc X,u}{\H_{h_p}}}{L_2(\Q)}\hspace{-0.25cm}< \infty
  \end{align}
  holds for all $u\in\H_{h_p}$, with a $u$-independent absolute constant in $\lesssim$.
\end{assumption}

\begin{example}[Applicability of Assumption~\ref{ass:sub-gaussian}] \label{example:assumption-dificult-verify} 
    In the simple case $d=1$, $k(x,y) = xy$ ($\H_k = \R$), and target measure $\P = \mathcal N(0,1)$, Assumption~\ref{ass:sub-gaussian} is satisfied, for instance, for $\Q = \operatorname{Unif}\left(-\sqrt 3,\sqrt 3\right)$. The details are as follows. From \eqref{eq:xi-p}, $\xi_p(\cdot,x) = h_p(\cdot,x) = 1-x^2$ ($x\in\R$). We note that $\E_{X\sim\Q}h_p(\cdot,X) = 0$ implies that $\bar h_p(\cdot,x) = h_p(\cdot, x)$ and thus we obtain $\norm{\ip{h_p(\cdot, X), u}{\R}}{\psi_2} = |u|\norm{1-X^2}{\psi_2} \stackrel{(a)}{\le} |u|c_1 \stackrel{(b)}{=} |u|c_1c_2 \norm{1-X^2}{L_2(\Q)} \lesssim \norm{\ip{h_p(\cdot, X), u}{\R}}{L_2(\Q)}$. The boundedness of $X$ implies the sub-Gaussianity (in the real-valued sense) of $1-X^2$ in (a); hence, $\norm{1-X^2}{\psi_2} \le c_1$. In (b), we let $c_2 = \norm{1-X^2}{L_2(\Q)}^{-1}$.
\end{example}

We elaborate further on Assumption~\ref{ass:sub-gaussian} in Remark~\ref{remark:main-thm}\ref{item:sub-gaussian-assumption}, after we state our following main result.

\begin{theorem}[Consistency of Nyström-KSD]
\label{thm:main-statement}
  Let Assumption~\ref{ass:sub-gaussian} hold, $C_{\Q,\bar h_p}\ne 0$, and assume the setting of Lemma~\ref{lemma:nystroem-ksd-estimator}. Then, for any $\delta \in (0,1)$ with $\left(\Q^n\otimes \Lambda^m\right)$-probability of at least $1-\delta$ it holds that
  \begin{align}
    \left|S_p(\Q) - \tilde S_p\left(\hat \Q_n\right)\right|  &\lesssim \frac{\sqrt{\trace\left(C_{\Q,\bar h_p}\right)}\log(6/\delta)}{n} 
     + \sqrt{\frac{\trace\left(C_{\Q,\bar h_p}\right)\log(6/\delta)}{n}}  \\
    &\quad +
\frac{\sqrt{\trace\left(C_{\Q,\bar h_p}\right)\log(12n/\delta)\log(12/\delta)}}{m} 
\sqrt{\mathcal N_{\Q,\bar h_p}\left(\frac{c\trace\left(C_{\Q,\bar h_p}\right)}{m}\right)}
  \end{align}
  when $  m\gtrsim \max\left\{\opnorm{C_{\Q,\bar h_p}}^{-1}\trace\left(C_{\Q,\bar h_p}\right),\log (12/\delta)\right\}$, where $c>1$ is a constant.

\end{theorem}

To interpret the consistency guarantee of Theorem~\ref{thm:main-statement}, we consider the three terms on the r.h.s.\ w.r.t.\ the magnitude of $m$. The first two terms converge with $\O\!\left(n^{-1/2}\right)$, independent of the choice of $m$.
By using the universal upper bound  $\mathcal N_{\Q,\bar h_p}\left(\frac{c\trace\left(C_{\Q,\bar h_p}\right)}{m}\right) \lesssim m$ on the effective dimension, the last term reveals that an overall rate of $\O\!\left(n^{-1/2}\right)$ can only be achieved with further assumptions regarding the rate of decay of the effective dimension if one also requires $m = o\left(n^{2/3}\right)$ --- as is necessary for a speed-up, see Remark~\ref{remark:nystroem-ksd-lemma}\ref{item:speedup}. Indeed, the rate of decay of the effective dimension can be linked to the rate of decay of the eigenvalues of the covariance operator \citep[Proposition~4,~5]{della21regularized}, which is known to frequently decay exponentially, or, at least, polynomially. In this sense, the last term acts as a balance, which takes the characteristics of the data and of the kernel into account.

The next corollary shows that an overall rate of $\O\left(n^{-1/2}\right)$ can be achieved, depending on the properties of the covariance operator.

\begin{corollary}\label{corr:decay-assumption} In the setting of Theorem~\ref{thm:main-statement}, assume that the spectrum of the covariance operator $C_{\Q,\bar h_p}$ decays either (i) polynomially, implying that $\mathcal N_{\Q,\bar h_p}(\lambda) \lesssim \lambda^{-\gamma}$ for some $\gamma\in(0,1]$, or
(ii) exponentially, implying that, $\mathcal N_{\Q,\bar h_p}(\lambda) \lesssim \log(1+\frac{c_1}{\lambda})$ for some $c_1>0$.
Then it holds that
  \begin{align}
    \left|S_p(\Q) - \tilde S_p\left(\hat \Q_n\right)\right| = \O_{\Q^n\otimes \Lambda^m}\!\left(\frac{1}{\sqrt n}\right),
  \end{align}
  assuming that the number of Nyström points satisfies
(i) $m\gtrsim n^{\frac{1}{2-\gamma}}\log^{\frac{1}{2-\gamma}}(12n/\delta)\log^{\frac{1}{2-\gamma}}(12/\delta)$ in the first case, or
(ii) $m\gtrsim \sqrt{n}\left(\log\left(1+\frac{c_1n}{c\trace\left(C_{\Q,\bar h_p}\right)}\right)\log(12n/\delta)\log(12/\delta)\right)^{1/2}$ in the second case.
\end{corollary}

To interpret these rates---see Remark~\ref{remark:main-thm}\ref{item:cmp-root-n-rates}---, we obtain the (matching) $\sqrt n$-consistency %
of the quadratic time estimator \eqref{eq:ksd-quad-time-v} in our following result.

\begin{theorem}[Consistency of KSD]\label{thm:v-stat-consistency}
  Assume that $\norm{\norm{\fm X}{\H_{h_p}}}{\psi_2}< \infty$ and define $\hat \Q_n = \{X_1,\ldots,X_n\}$, where $\{X_i\}_{i\in[n]}\stackrel{\text{i.i.d.}}{\sim} \Q$. Then it holds that
  \begin{align}
      \left|S_p(\Q) - S_p\left( \hat \Q_n \right)\right| = \mathcal O_{\Q^n}\left(\frac{1}{\sqrt n}\right).
  \end{align}
\end{theorem}

The following example illustrates that, in some cases, the assumption $\norm{\norm{\fm X}{\H_{h_p}}}{\psi_2}< \infty$ can be verified analytically.

\begin{example}[Assumption $\norm{\norm{\fm X}{\H_{h_p}}}{\psi_2}< \infty$] \label{example:assumption-easy-verify} Assume that $d=1$, $k=\exp\left(-\gamma(x-y)^2\right)$ ($\gamma>0$), target measure $\P = \mathcal N(0,1)$, and samples $X,X_1,\ldots,X_n \stackrel{\text{i.i.d.}}{\sim}\Q$ with $\norm{X}{\psi_2}< \infty$. Then
\begin{align}
    \norm{\norm{h_p(\cdot,X)}{\H_{h_p}}}{\psi_2}^2 &\stackrel{(a)}{=} \norm{h_p(X,X)}{\psi_1} \stackrel{(b)}{=} \norm{X^2+2\gamma}{\psi_1} \notag  \\
    &\stackrel{(c)}{\le} \norm{X^2}{\psi_1} + \norm{2\gamma}{\psi_2} \stackrel{(d)}{=} \norm{X}{\psi_2} + \frac{2\gamma}{\sqrt {\log 2}} < \infty, \notag
\end{align}
with the following details. Lemma~\ref{lemma:orlicz-properties}(iv) and the reproducing property yield (a). (b) follows from the explicit form of $h_p$ given in Example~\ref{example:unbounded-kernel}(i).  The triangle inequality gives (c) and (d) follows from the definition of the $\psi_2$-norm using that $2\gamma$ is non-random.

In this setting, similar computations using Example~\ref{example:unbounded-kernel}(ii) show that the assumption is also satisfied with the IMQ kernel.
\end{example}

A few remarks are in order.

\begin{remark}\label{remark:main-thm}~
  \begin{enumerate}[label=(\alph*)]
    \item \tb{Runtime benefit.} Recall that --- see Remark~\ref{remark:nystroem-ksd-lemma}\ref{item:speedup} ---, our proposed Nyström estimator \eqref{eq:nystroem-ksd-estimator} requires $m = o\left(n^{2/3}\right)$ Nyström samples to achieve a speed-up. Hence, in the case of polynomial decay, an asymptotic speed-up with a statistical accuracy that matches the quadratic time estimator \eqref{eq:ksd-quad-time-v} is guaranteed for $\gamma < 1/2$; in the case of exponential decay, large enough $n$ always suffices.

    \item \tb{Comparison of Theorem~\ref{thm:chatalic} and Theorem~\ref{thm:main-statement}.} \label{item:cmp-to-chatalic} Recall that both theorems target precisely the same estimators, \citet[(5)]{chatalic22nystrom} and \eqref{eq:nystroem-ksd-estimator}, respectively. We note that in the finite-dimensional case, every bounded random variable is also sub-Gaussian. This property does not carry over to sub-Gaussianity in the infinite-dimensional case; see the remark after \citet[Definition 1]{della21regularized}. In this sense, the assumptions of both statements are not directly comparable. Still, the takeaway of both results---with these different sets of conditions---is the same.

    \item \tb{Sub-Gaussian assumption.} \label{item:sub-gaussian-assumption} Key to the proof of Theorem~\ref{thm:main-statement} is having an adequate notion of non-boundedness of the feature map. One approach---common for controlling unbounded real-valued random variables--- is to impose a sub-Gaussian assumption. In Hilbert spaces, various definitions of sub-Gaussian behavior have been investigated \citep{talagrand87subgauss,fukuda1990subgauss,antonini97subgaussian}; see \citet{giorgobiani20notessubgauss} for a recent survey. Among the definitions of sub-Gaussianity, we carefully selected \citet[Def.~2]{koltchinskii17concentration}.\footnote{The condition is also referred to as \textit{sub-Gaussian in Fukuda's sense} \citep[Def.~1]{giorgobiani20notessubgauss}.} Specifically, this assumption allows us to derive our key Lemma~\ref{lemma:bound-projection} and Lemma~\ref{lemma:sub-gauss-norm}. The former is similar to \citet[Lemma~6]{rudi15less}, which is typically employed for Nyström analysis in the bounded case \citep{chatalic22nystrom,sterge22nystroem,kalinke23nystrom}, but our result applies to the sub-Gaussian setting.
    The main technical challenge we resolve is transforming our setting to a form in which existing concentration results can be leveraged. Especially the case of $\P \neq \Q$ requires special care, which we tackle by systematically using the centered covariance operator $C_{\Q,\bar h_p}$; we refer to the respective proof for details.\footnote{We note that an analysis of the centered setting is also challenging in the bounded case; for instance, \citet{sterge22nystroem} tackle the resulting difficulties (in case of kernel PCA) with U-statistics, of which our method is independent.}
    The latter, Lemma~\ref{lemma:sub-gauss-norm}, intuitively states that norms of sub-Gaussian vectors whitened by $C_{\Q,\bar h_p,\lambda}^{-1/2}$ inherit the sub-Gaussian property.
    Together, these lemmas open the door to proving Theorem~\ref{thm:main-statement}.

    \item \tb{Comparison of Theorem~\ref{thm:main-statement} and Theorem~\ref{thm:v-stat-consistency}.} \label{item:cmp-root-n-rates} With the weaker RKHS norm condition $\norm{\norm{\fm X}{\H_{h_p}}}{\psi_2}<\infty$ (implied by Assumption~\ref{ass:sub-gaussian}, see Lemma~\ref{lemma:sub-gauss-norm}), Theorem~\ref{thm:v-stat-consistency} shows that the quadratic time estimator \eqref{eq:ksd-quad-time-v} converges with rate $\O\!\left( n^{-1/2} \right)$. Our Nyström result, Theorem~\ref{thm:main-statement} with Corollary~\ref{corr:decay-assumption}, shows that a matching rate can be achieved (given an appropriate decay of the effective dimension) with $m=\tilde \Theta\!\left( \sqrt n \right)$; this choice of $m$ satisfies $m=o\left( n^{2/3} \right)$ and thus implies an asymptotic speedup by~(a).

    \item \tb{General KSD framework.} We note that our results also hold in the general KSD framework \citep{hagrass24stein} but we present them on $\R^d$, which one arguably most frequently encounters in practice, to simplify exposition.

  \end{enumerate}
\end{remark}

\section{EXPERIMENTS}
\label{sec:experiments}

\begin{figure*}
  \centering
  \includegraphics[width=1\textwidth,trim=0 22pt 0 19pt]{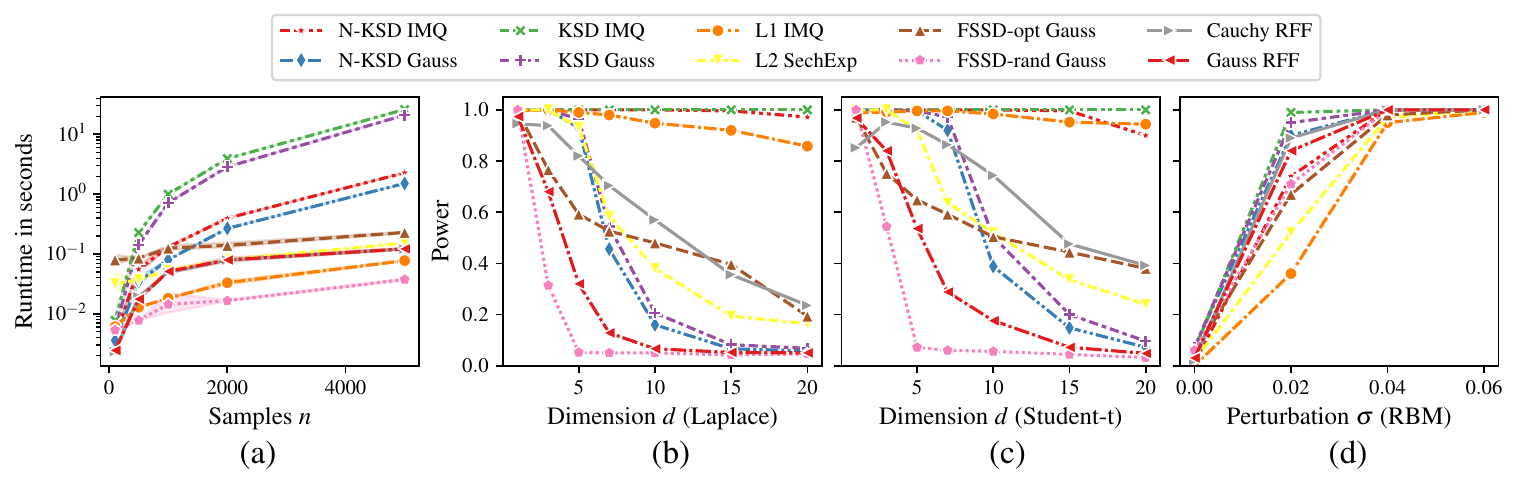}
  \caption{Comparison of goodness-of-fit tests w.r.t.\ their runtime and their power.}
  \label{fig:exp-1}
\end{figure*}

We verify the viability of our proposed method, abbreviated as N-KSD in this section, by comparing its runtime and its power to existing methods: the quadratic time KSD \citep{liu16kernelized,chwialkowski16kernel}, the linear-time goodness-of-fit test finite set Stein discrepancy (FSSD; \citealt{jitkrittum17lineartime}), RFF-based KSD approximations \citep{huggins18random}, and the linear-time goodness-of-fit test using random feature Stein discrepancy (L1 IMQ, L2 SechExp; \citealt{huggins18random}).\footnote{\label{fn:code}The code replicating our experiments is available at \url{https://github.com/FlopsKa/nystroem-ksd}.} For FSSD, we consider randomized test locations (FSSD-rand) and optimized test locations (FSSD-opt); the optimality is meant w.r.t.\ a power proxy detailed in the cited work. %
For all competitors, we use the settings and implementations provided by the respective authors.
We use the well-known Gaussian kernel $k(\b x, \b y) = \exp\left(-\gamma\norm{\b x- \b y}{\R^d}^2\right)$ ($\gamma > 0$) with the median heuristic \citep{garreau18large}, and the IMQ kernel $k(\b x, \b y) = \left(c^2+\norm{\b x - \b y}{\R^d}^2\right)^{-\beta}$ \citep{gorham17measuring}, with the choices of $\beta, c >0$ detailed in the respective experiment description. To approximate the null distribution of N-KSD, we perform a bootstrap with \eqref{eq:nys-bootstrap}, setting $D=500$. To allow an easy comparison, our experiments replicate goodness-of-fit testing experiments from \citet{chwialkowski16kernel,jitkrittum17lineartime} and \citet{huggins18random}. For additional results, we refer to Appendix~\ref{sec:additional-results}. We ran all experiments on a PC with Ubuntu 20.04, 124GB RAM, and 32 cores with 2GHz each.

\tb{Runtime.} We set $m=4\sqrt n$ for N-KSD to match the settings in our other experiments.
    As per recommendation, we fix the number of test locations $J=10$ for L1 IMQ, L2 SechExp, Cauchy RFF, Gauss RFF, and both FSSD methods. The data is randomly generated with $d=10$ dimensions. We note that the dimensionality enters the complexity only through the kernel evaluation; the dependence is linear in our case.
The runtime, see Figure~\ref{fig:exp-1}(a) for the average over 10 repetitions (the error bars indicate the estimated $95\%$ quantile), behaves as predicted by the complexity analysis.
The proposed approach runs orders of magnitudes faster than the quadratic time KSD estimator \eqref{eq:ksd-quad-time-v}.
From $n=1500$, all (near-)linear-time approaches are faster (excluding FSSD-opt, which has a relatively large fixed cost). Still, N-KSD achieves competitive runtime results even for $n=5000$.

\tb{Laplace vs.\ standard normal.} We fix the target distribution $\P = \mathcal N\left(\bm 0_d,\b I_d\right)$ and obtain $n=1000$ samples from the alternative $\Q = \operatorname{Lap}\left(0,\frac{1}{\sqrt 2}\right)^d$, a product of $d$ Laplace distributions. We test $H_0 : \Q = \P$ vs.\ $H_1 : \Q \neq \P$ with a level of $\alpha=0.05$. We set the kernel parameters $c$ and $\beta$ for KSD IMQ and N-KSD IMQ as per the recommendation for L1 IMQ in the corresponding experiment by \citet{huggins18random}.
Figure~\ref{fig:exp-1}(b) reports the power (obtained over $500$ draws of the data) of the different approaches. KSD Gauss and its approximation N-KSD Gauss perform similarly but their power diminishes from $d=3$. KSD IMQ achieves full power for all tested dimensions and performs best overall. N-KSD IMQ ($m=4\sqrt n$) achieves comparable results, with a small decline from $D=15$. Our proposed method outperforms all existing KSD accelerations.

\tb{Student-t vs.\ standard normal.} The setup is similar to that of the previous experiment, but we consider samples from $\Q$ a multivariate student-t distribution with $5$ degrees of freedom, set $n=2000$, and repeat the experiment 250 times to estimate the power.
We show the results in Figure~\ref{fig:exp-1}(c).
All approaches employing the Gaussian kernel quickly loose in power; all techniques utilizing the IMQ kernel, including N-KSD IMQ, achieve comparably high power throughout.

\tb{Restricted Boltzmann machine (RBM).} Similar to \citet{liu16kernelized,jitkrittum17lineartime}, we consider the case where the target $\P$ is the non-normalized density of an RBM with $50$ visible and $40$ hidden dimensions; the samples $\hat \Q_n$ are obtained from the same RBM perturbed by independent Gaussian noise with variance $\sigma^2$. For $\sigma^2 = 0$, $H_0 : \Q = \P$ holds, and for $\sigma^2 >0$, implying that the alternative $H_1 : \Q \neq \P$ holds, the goal is to detect that the $n=1000$ samples come from a forged RBM. For the IMQ kernel (L1 IMQ, N-KSD IMQ, KSD IMQ), we set $c=1$ and $\beta=-1/2$. We show the results in Figure~\ref{fig:exp-1}(d), using $100$ repetitions to obtain the power. KSD with the IMQ and with the Gaussian kernel performs best. Our proposed Nyström-based method ($m=4\sqrt n$) nearly matches its performance with the IMQ kernel while requiring only a fraction of the runtime. Besides Cauchy RFF and Gauss RFF, all other approaches achieve less power for $\sigma \in\{0.02,0.04\}$.

These experiments demonstrate the efficiency of the proposed Nyström-KSD method.

\section{LIMITATIONS} \label{sec:limitations}
Assumption~\ref{ass:sub-gaussian}, which underpins our main result (Theorem~\ref{thm:main-statement}), can be difficult to verify in some cases. We refer to Example~\ref{example:assumption-dificult-verify} for a case where the analytical verification is possible. The weaker assumption   $\norm{\norm{\fm X}{\H_{h_p}}}{\psi_2}< \infty$ of Theorem~\ref{thm:v-stat-consistency} is usually easier to verify analytically, as we show in Example~\ref{example:assumption-easy-verify}.
We note that, as with all kernel-based tests, the choice of the kernel, corresponding to the setting of $\gamma$ for the Gaussian kernel (resp.\ the setting of $\beta, c$ for the IMQ kernel), has an impact on the power of the test. While optimizing kernel parameters is not the focus of this work, there exist methods in the literature to (approximately) achieve this goal \citep{jitkrittum16interpretable,jitkrittum17adaptive2,jitkrittum17lineartime,liu20learning,schrab22ksdagg,schrab22efficient,hagrass22spectral,hagrass23spectralgof,hagrass24stein}.

\subsubsection*{Acknowledgements}
This work was supported by the pilot program Core-Informatics of the Helmholtz Association~(HGF). BKS is partially supported by the National Science Foundation (NSF) CAREER award DMS-1945396.

\bibliography{bib/collected_Zoltan.bib,bib/collected_plus.bib,bib/publications.bib}

\newpage
\appendix
\setcounter{equation}{0}
\renewcommand{\theequation}{\thesection.\arabic{equation}}

\section{PROOFS}\label{sec:proofs}
This section is dedicated to the proofs of our results in the main text.
Lemma~\ref{lemma:nystroem-ksd-estimator} is proved in Section~\ref{sec:proof-lemma-nys-ksd}. We prove our main result (Theorem~\ref{thm:main-statement}) in Section~\ref{sec:proof-main-stmt}; Corollary~\ref{corr:decay-assumption} is shown in Section~\ref{sec:proof-corollary}. The proof of Theorem~\ref{thm:v-stat-consistency} is in Section~\ref{sec:proof-v-stat-consistency}.

\subsection{Proof of Lemma~\ref{lemma:nystroem-ksd-estimator}}\label{sec:proof-lemma-nys-ksd}
By \eqref{eq:stein-discrepancy}, KSD is the norm of the mean embedding of $\Q$ under $\fm{\cdot}$, that is,
\begin{align}
  \label{eq:mu-ksd}
  S_p(\Q) = \norm{\int_{\R^d}\fm{\b x}\d\Q(\b x)}{\H_{h_p}} = \norm{\mu_{h_p}(\Q)}{\H_{h_p}} .
\end{align}
Hence, with \citet[(5)]{chatalic22nystrom}, the optimization problem \eqref{eq:optim-projection} has the solution $\bm\alpha = (\alpha_i)_{i=1}^m = \frac1n\b K_{h_p,m,m}^- \b K_{h_p,m,n}\bm1_n \in \R^m$. Now, using \eqref{eq:mu-ksd}, we have
\begin{align*}
  \MoveEqLeft\norm{\sum_{i=1}^m\alpha_i \fm{\tilde {\b x}_i}}{\H_{h_p,m}}^2
  \stackrel{(a)}{=} \left \langle \sum_{i=1}^m\alpha_i\fm{\tilde {\b x}_i}, \sum_{i=1}^m\alpha_i\fm{\tilde {\b x}_i} \right\rangle_{\H_{h_p,m}}\\
  &\stackrel{(b)}{=} \sum_{i=1}^m \sum_{j=1}^m \alpha_i \alpha_j\left \langle \fm{\tilde {\b x}_i}, \fm{\tilde {\b x}_j} \right\rangle_{\H_{h_p,m}} \stackrel{(c)}{=} \sum_{i=1}^m\sum_{j=1}^m\alpha_i\alpha_jh_p\left(\tilde {\b x}_i,\tilde {\b x}_j\right)
  \stackrel{(d)}{=} \bm\alpha\T \b K_{h_p,m,m}\bm\alpha \\
  & \stackrel{(e)}{=} \frac{1}{n^2}\bm1_n\T\b K_{h_p,n,m}\underbrace{\b K_{h_p,m,m}^-\b K_{h_p,m,m}\b K_{h_p,m,m}^-}_{=\b K_{h_p,m,m}^-}\b K_{h_p,m,n}\bm1_n = \bm \beta_p\T\b K_{h_p,m,m}^-\bm\beta_p.
\end{align*}
In (a) we used that $\left\|\cdot\right\|_{\H_{h_p,m}}$ is inner product induced, (b) follows from the linearity of the inner product, (c) is implied by the reproducing property, (d) is by the definition of the Gram matrix, in (e) we made use of the explicit form of $\bm \alpha$, the symmetry of
Gram matrices, the property $\b K_{h_p,m,n}\T = \b K_{h_p,n,m}$, and that the Moore-Penrose inverse satisfies $\b A^-\b A \b A^- = \b A^-$ for any matrix $\b A$.

\subsection{Proof of Theorem~\ref{thm:main-statement}}\label{sec:proof-main-stmt}

Contrasting the existing related work \citep{rudi15less,chatalic22nystrom,sterge22nystroem,kalinke23nystrom}, we do not impose a boundedness assumption on the feature map. This relaxation leads to new technical difficulties that we resolve in the following. We start our analysis from a decomposition similar to \citet[Lemma~4.1]{chatalic22nystrom}; the difference is that we introduce the centered covariance operator $C_{\Q,\bar h_p,\lambda}$ which allows us to handle both $\P=\Q$ and the challenging case of $\P \ne \Q$ in a unified fashion.

To simplify notation, let $\mu_{h_p} := \mu_{h_p}(\Q)$, $\hat \mu_{h_p} := \mu_{h_p}\left(\hat \Q_n\right)$, and $\hat \mu_{h_p}^\nys := P_{\H_{h_p,m}}\mu_{h_p}\left(\hat \Q_n\right)$. First, we decompose the error as follows.\newpage %
\begin{align}
  \MoveEqLeft\left|S_p(\Q) - \tilde S_p\left(\hat \Q_n\right)\right|
  \stackrel{(a)}{=} \left|\norm{\mu_{h_p}}{\H_{h_p}}- \norm{\hat \mu_{h_p}^\nys}{\H_{h_p}}\right|
  \stackrel{(b)}{\le} \norm{\mu_{h_p}-\hat\mu_{h_p}^\nys} {\H_{h_p}}  \\
  &\stackrel{(c)}{\le} \norm{\mu_{h_p}-\hat\mu_{h_p}} {\H_{h_p}} + \norm{\hat \mu_{h_p}-\hat\mu_{h_p}^\nys} {\H_{h_p}} \\
  &\stackrel{(d)}{=} \norm{\mu_{h_p}-\hat\mu_{h_p}} {\H_{h_p}} + \norm{\left(I-P_{\H_{h_p,m}}\right)\left(\hat \mu_{h_p}-\frac{1}{m}\sum_{i=1}^{m}\fm{\tilde{\b x}_i}\right)} {\H_{h_p}} \\
  &\stackrel{(e)}{\le} \underbrace{\norm{\mu_{h_p}-\hat\mu_{h_p}} {\H_{h_p}}}_{=:t_1} + \underbrace{\opnorm{\left(I-P_{\H_{h_p,m}}\right)C_{\Q,\bar h_p,\lambda}^{1/2}}}_{=:t_2} \underbrace{\norm{C_{\Q,\bar h_p,\lambda}^{-1/2}\left(\hat \mu_{h_p}-\frac{1}{m}\sum_{i=1}^{m}\fm{\tilde{\b x}_i}\right)} {\H_{h_p}}}_{=:t_3}. \label{eq:decomposition}
\end{align}
(a) is implied by \eqref{eq:mu-ksd} and \eqref{eq:KSD-estimator};
(b) follows from the reverse triangle inequality;
$\pm \hat\mu_{h_p}$ and the triangle inequality yield (c);
in (d), we use the distributive property and introduce the term $\frac1m\sum_{i=1}^{m}\fm{\tilde {\b x}_i} \in \H_{h_p,m}$ whose projection onto the orthogonal complement of $\H_{h_p,m}$  is zero; in (e) $I = C_{\Q,\bar h_p,\lambda}^{1/2}C_{\Q,\bar h_p,\lambda}^{-1/2}$ was introduced and we used the definition of the operator norm.

We next obtain individual probabilistic bounds for the three terms $t_1$, $t_2$, and $t_3$, which we subsequently combine by union bound. We will then conclude the proof by showing that all assumptions that we imposed along the way are satisfied.

\begin{itemize}
  \item \tb{Term $t_1$.} The first term measures the deviation of an empirical
  mean $\hat\mu_{h_p}$ to its population counterpart $\mu_{h_p}$. To bound this deviation $\norm{\hat\mu_{h_p}-\mu_{h_p}} {\H_{h_p}}=\left\|\frac{1}{n}\sum_{i=1}^n \fmc{\b x_i}\right\|_{\H_{h_p}}$, we will use the Bernstein inequality (Theorem~\ref{thm:bernstein-hilbert}) with the $\eta_i := \fmc{\b x_i}\in \H_{h_p}$ $(i\in [n])$ centered random variables, by gaining control on the moments of $Y:=\left\|\fmc X\right\|_{\H_{h_p}}$. This is what we elaborate next.

  By Assumption~\ref{ass:sub-gaussian} and Lemma~\ref{lemma:sub-gauss-norm}, $Y$ is sub-Gaussian; hence it is also sub-exponential (Lemma~\ref{lemma:orlicz-properties}(3)), and therefore (Lemma~\ref{lemma:sub-exp-bernstein}) it satisfies the Bernstein  condition
  \begin{align}
  \E|Y|^p &\le \frac{1}{2}p!\sigma^2B^{p-2} < \infty, \quad\text{with}&
   \sigma &= \sqrt 2 \norm{Y}{\psi_1} ,&  B &= \norm{Y}{\psi_1},
  \end{align}
  for any $p\ge 2$. Notice that $B = \norm{Y}{\psi_1} \stackrel{(a)}{\lesssim}\norm{Y}
   {\psi_2} \stackrel{(b)}{\lesssim} \sqrt{\trace\left(C_{\Q,\bar h_p}\right)}$.
  (a) follows from  Lemma~\ref{lemma:orlicz-properties}(3) and (b) is implied by Lemma~\ref{lemma:sub-gauss-norm}. As $\sigma \asymp B$, we also got that $\sigma \lesssim \sqrt{\trace\left(C_{\Q, \bar h_p}\right)}$.

  Having obtained a bound on the moments, we can apply Bernstein's inequality for separable Hilbert spaces (\citealt{yurinsky95sums}; recalled in Theorem~\ref{thm:bernstein-hilbert}) to the centered
  $\eta_i = \fmc{\b x_i} \in \H_{h_p}$-s ($i\in[n]$), and obtain that for any $\delta \in (0,1)$ it holds that
  \begin{align}
    \Q^n\Bigg(\underbrace{\norm{\mu_{h_p}-\hat\mu_{h_p}}{\H_{h_p}}}_{(= \norm{\frac1n\sum_{i=1}^{n} \eta_i}{\H_{h_p}})} \hspace{-0.2cm} \lesssim \frac{\sqrt{\trace\left(C_{\Q,\bar h_p}\right)}\log(6/\delta)}{n} + \sqrt{\frac{\trace\left(C_{\Q,\bar h_p}\right)\log(6/\delta)}{n}}\Bigg) \ge 1-\delta/3.\quad\quad \label{eq:t1-bound}
  \end{align}
  Note that \eqref{eq:t1-bound} also holds with the measure $\Q^n\otimes \Lambda^m$, since the event considered in \eqref{eq:t1-bound} has no randomness w.r.t.\ $\Lambda^m$.

  \item \tb{Term $t_2$.} Assume that $0 < \lambda \le \opnorm{C_{\Q,\bar h_p}}$. Then, we can handle the second term with Lemma~\ref{lemma:bound-projection} and obtain that for any $\delta\in(0,1)$ it holds that
  \begin{align}
    \left(\Q^n\otimes \Lambda^m\right)\left(\opnorm{\left( I-P_{\H_{h_{p},m}} \right)C_{\Q,\bar h_{p},\lambda}^{1/2}} \lesssim \sqrt \lambda\right) \ge 1-\delta/3 \label{eq:t2-bound}
  \end{align}
  provided that $m\gtrsim \max\left\{ \frac{\trace\left(C_{\Q,\bar h_p}\right)}{\lambda},1  \right\}\log\left(12/\delta\right)$.

  \item \tb{Term $t_3$.} The third term depends on the  sample $(\b x_i)_{i=1}^n \stackrel{\text{i.i.d.}}{\sim}\Q$ and on the Nyström selection $(i_j)_{j=1}^m \stackrel{\text{i.i.d.}}{\sim} \text{Unif}([n]) =: \Lambda$; with this notation $\tilde{\b x}_j = \b x_{i_j}$ ($j\in [m]$). Our goal is to take both sources of randomness into account.

\tb{Fixed $\b x_i$-s, randomness in $i_j$-s:}
      Let the sample $(\b x_i)_{i=1}^n$ be fixed. As the $(\b x_{i_j})_{j=1}^m$-s are i.i.d.,
  \begin{align}
  t_3&=\Bigg\|C_{\Q,\bar h_p,\lambda}^{-1/2}\left(\hat \mu_{h_p}-\frac{1}{m}\sum_{i=1}^{m}\fm{\tilde{\b x}_i}\right)\Bigg\|_{\H_{h_p}} = \Bigg\|\frac{1}{m}\sum_{i=1}^{m} \underbrace{\left[C_{\Q, \bar h_p,\lambda}^{-1/2}\left(\fm{\tilde{\b x}_i}-\hat \mu_{h_p}\right)\right]}_{=:Y_i}\Bigg\|_{\H_{h_p}}
  \end{align}
  measures the concentration of the sum $\frac{1}{m}\sum_{i=1}^mY_i$ around its expectation, which is zero as $\E_J\left[h_p(\cdot,\b x_J)\right] = \hat{\mu}_{h_p}$ with $J \sim \Lambda$. Notice that
  \begin{align}
  \MoveEqLeft \max_{i\in[m]}\norm{Y_i}{\H_{h_p}} =
  \max_{i\in[m]}\norm{C_{\Q,\bar h_p,\lambda}^{-1/2}\left(\fm{\tilde{\b x}_i}-\hat \mu_{h_p}\right)}{\H_{h_p}} \\
  &=\max_{i\in[m]}\norm{C_{\Q,\bar h_p,\lambda}^{-1/2}\left(\fm{\tilde{\b x}_i} - \E_{X\sim \Q} \fm X + \E_{X\sim \Q} \fm X -\hat \mu_{h_p}\right)}{\H_{h_p}} \\
  &\le\max_{i\in[m]}\Big\|C_{\Q,\bar h_p,\lambda}^{-1/2}\big(\underbrace{\fm{\tilde{\b x}_i} - \E_{X\sim \Q} \fm X}_{= \fmc{\tilde{\b x}_i}}\big)\Big\|_{\H_{h_p}} \hspace{-0.2cm}+ \norm{C_{\Q,\bar h_p,\lambda}^{-1/2}\left(\hat \mu_{h_p} - \E_{X\sim \Q} \fm X\right)}{\H_{h_p}} \\
  &\le\max_{i\in[n]}\norm{C_{\Q,\bar h_p,\lambda}^{-1/2}\fmc{\b x_i}}{\H_{h_p}} + \norm{C_{\Q,\bar h_p,\lambda}^{-1/2}\left(\hat \mu_{h_p} - \E_{X\sim \Q} \fm X\right)}{\H_{h_p}} \\
  &\le\max_{i\in[n]}\norm{C_{\Q,\bar h_p,\lambda}^{-1/2}\fmc{\b x_i}}{\H_{h_p}} + \frac1n\sum_{i\in[n]}\Big\|{C_{\Q,\bar h_p,\lambda}^{-1/2}\big(\underbrace{\fm {\b x_i} - \E_{X\sim \Q} \fm X}_{= \fmc{\b x_i}}\big)}\Big\|_{\H_{h_p}} \\
  &\le 2 \max_{i\in[n]}\norm{C_{\Q,\bar h_p,\lambda}^{-1/2}\fmc{\b x_{i}}}{\H_{h_p}} =: K = K(\b x_1,\ldots,\b x_n),
  \end{align}
   where we used that $\pm \E_{X\sim \Q}\fm X = 0$, the triangle inequality, and the homogeneity of the norm. An application of Theorem~\ref{thm:bernstein-bounded} yields that, conditioned on the sample $(\b x_i)_{i=1}^n$, it holds that
  \begin{align}
    \Lambda^m \!\left( (i_j)_{j=1}^m \,:\, t_3 \le K\frac{\sqrt{2\log(12/\delta)}}{\sqrt m} \enspace \Bigg|\,\left(\b x_i\right)_{i=1}^n\right) \ge 1-\frac{\delta}{6}. \label{eq:t3-first-bound}
  \end{align}

 \tb{Randomness in $\b x_i$-s:}
  Let $Z_i:=\norm{C_{\Q, \bar h_p,\lambda}^{-1/2}\fmc{\b x_i}}{\H_{h_p}}$ ($i\in[n]$) with  $(\b x_i)_{i=1}^n \stackrel{\text{i.i.d.}}\sim\Q$. By Assumption~\ref{ass:sub-gaussian} and Lemma~\ref{lemma:sub-gauss-norm}, the $Z_i$-s are sub-Gaussian random variables. Hence, by Lemma~\ref{lemma:max-of-sub-gauss}, with probability at least $1-\delta/6$, it holds that
  \begin{align}
      K = 2\max_{i\in[n]}|Z_i|\lesssim \sqrt{\norm{Z_1}{\psi_2}^2\log(12n/\delta)}. \label{eq:first-max}
  \end{align}

  By Lemma~\ref{lemma:sub-gauss-norm}, $\norm{Z_1}{\psi_2}^2 \lesssim \trace\left(C_{\Q,\bar h_p,\lambda}^{-1}C_{\Q,\bar h_p}\right)$.
  We have shown that
  \begin{align}
      \Q^n\!\left( \left(\b x_i\right)_{i=1}^n \,:\, K\lesssim \sqrt{\trace\left(C_{\Q, \bar h_p,\lambda}^{-1}C_{\Q,\bar h_p}\right)\log(12n/\delta)}\right) \ge 1-\frac{\delta}{6}. \label{eq:t3-second-bound}
  \end{align}

  \tb{Combination:} We now combine the intermediate results. Let
  \begin{align}
      A &= \left\{\left(\left(\b x_i\right)_{i=1}^n,\left(i_j\right)_{j=1}^m\right) \;:\; t_3 \lesssim \frac{\sqrt{\trace\left(C_{\Q, \bar h_p,\lambda}^{-1}C_{\Q,\bar h_p}\right) \log(12n/\delta)\log(12/\delta)}}{\sqrt m}\right\}, \\
      B &= \left\{ \left(\b x_i\right)_{i=1}^n \,:\, K\lesssim \sqrt{\trace\left(C_{\Q, \bar h_p,\lambda}^{-1}C_{\Q,\bar h_p}\right)\log(12n/\delta)}\right\}, \\
      C &= \left\{ \left(\left(\b x_i\right)_{i=1}^n,\left(i_j\right)_{j=1}^m\right) \;:\;t_3 \le K\frac{\sqrt{2\log(12/\delta)}}{\sqrt m}, \left(\b x_i\right)_{i=1}^n \in B\right\} \subseteq A.
  \end{align}
  Then, with $\Q^n\otimes \Lambda^m$ denoting the product measure of $\Q^n$ and $\Lambda^m$, we obtain
  \begin{align}
      \MoveEqLeft\left(\Q^n\otimes \Lambda^m\right)\!\left( A \right) = \E_{\Q^n}\!\left[\Lambda^m\left(A \mid \xis \right)\right] = \int_{\left(\R^d\right)^n} \Lambda^m\!\left(A \mid \xis \right)\d \Q^n(\b x_1,\ldots, \b x_n) \\
      &\ge \int_B \Lambda^m\!\left(A \mid \xis \right)\d \Q^n(\b x_1,\ldots, \b x_n) \ge \int_B \Lambda^m\!\left(C \mid \xis \right)\d \Q^n(\b x_1,\ldots, \b x_n)\\
      &\stackrel{(a)}\ge \left(1-\frac{\delta}{6}\right) \Q^n(B) \stackrel{(b)}{\ge} (1-\delta/6)^2=1-\delta/3+\delta^2/6^2>1-\delta/3. \label{eq:t3-bound}
  \end{align}
  (a) is implied by the uniform lower bound established in \eqref{eq:t3-first-bound}. (b) was shown in \eqref{eq:t3-second-bound}.
\end{itemize}

  \paragraph{Combination of $t_1$, $t_2$, and $t_3$.} To conclude, we use decomposition \eqref{eq:decomposition}, and union bound \eqref{eq:t1-bound}, \eqref{eq:t2-bound}, and \eqref{eq:t3-bound}. Further, we observe that $\trace\left(C_{\Q, \bar h_p,\lambda}^{-1}C_{\Q,\bar h_p}\right) = \mathcal N_{\Q,\bar h_p}(\lambda)$, and obtain that
  \begin{align}
    \left(\Q^n\otimes \Lambda^m\right)\Bigg(\left|S_p(\Q) - \tilde S_p\left(\hat \Q_n\right)\right|  &\lesssim
    \frac{\sqrt{\trace\left(C_{\Q, \bar h_p}\right)}\log(6/\delta)}{n} + \sqrt{\frac{\trace\left(C_{\Q,\bar h_p}\right)\log(6/\delta)}{n}} + \\
    &\quad +
    \sqrt{\frac{\lambda\mathcal N_{\Q,\bar h_p}(\lambda)\log(12n/\delta)\log(12/\delta)}{m}}\Bigg) \ge 1-\delta \\
      \label{eq:main-thm-overall}
  \end{align}
  provided that $m\gtrsim \max\left\{\frac{\trace\left(C_{\Q,\bar h_p}\right)}{\lambda},1\right\}\log (12/\delta)$ and $0 < \lambda \le \opnorm{C_{\Q,\bar h_p}}$ both hold. Now, specializing $\lambda = \frac{c\trace\left(C_{\Q,\bar h_p}\right)}{m}$ for some absolute constant $c>1$, all constraints are satisfied for $m\gtrsim\max\left\{\log(12/\delta), \trace\left(C_{\Q,\bar h_p}\right)\opnorm{C_{\Q,\bar h_p}}^{-1}\right\}$. Using our choice of $\lambda$, after rearranging, we get the stated claim.

\subsection{Proof of Corollary~\ref{corr:decay-assumption}}\label{sec:proof-corollary}

The proof is split into two parts. The first one considers the polynomial decay assumption, the second one is about the exponential decay assumption.

\begin{itemize}
  \item \tb{Polynomial decay.} The $\sqrt n$-consistency of the first two addends in Theorem~\ref{thm:main-statement} was established in the discussion following the statement. Hence, we limit our considerations to the last addend. Assume that $\mathcal N_{\Q,\bar h_p}(\lambda) \lesssim \lambda^{-\gamma}$ for $\gamma \in (0,1]$. Observing that the trace expression is constant, the last addend in Theorem~\ref{thm:main-statement} yields that
  \begin{align}
    \sqrt{\frac{\log(12/\delta)\log(12n/\delta)\mathcal N_{\Q,h_p}\left(\frac{c\trace\left(C_{\Q,\bar h_p}\right)}{m}\right)}{m^2}}
    \stackrel{(a)}{\lesssim}\sqrt{\frac{\log(12/\delta)\log(12n/\delta)}{m^{2-\gamma}}} \stackrel{(b)}{=} \O\left(\frac{1}{\sqrt n}\right),
    \label{eq:poly-decay-second-addend}
  \end{align}
  with (a) implied by the polynomial decay assumption and (b) follows from our choice of $m\gtrsim n^{\frac{1}{2-\gamma}}\log^{\frac{1}{2-\gamma}}(12n/\delta)\log^{\frac{1}{2-\gamma}}(12/\delta)$.  This derivation confirms the first stated result.

  \item \tb{Exponential decay.} Assume it holds that $\mathcal N_{\Q,\bar h_p}(\lambda) \lesssim \log(1+c_1/\lambda)$. Observe that as per the discussion following Theorem~\ref{thm:main-statement}, the first two addends are $\O\left(n^{-1/2}\right)$. For the last addend, again noticing that the trace is constant, we have
  \begin{align}
  \MoveEqLeft\sqrt{\frac{\log(12/\delta)\log(12n/\delta)\mathcal N_{\Q,\bar h_p}\!\left(\frac{c\trace\left(C_{\Q,\bar h_p}\right)}{m}\right)}{m^2}}\hspace{-0.1cm}
    \stackrel{(a)}{\lesssim}\hspace{-0.1cm} \sqrt{\frac{\log(12/\delta)\log(12n/\delta)\log\left(1+\frac{c_1m}{c\trace\left(C_{\Q,\bar h_p}\right)}\right)}{m^2}} \\
    &\stackrel{(b)}{\lesssim} \sqrt{\frac{\log(12/\delta)\log(12n/\delta)\log\left(1+\frac{c_1n}{c\trace\left(C_{\Q,\bar h_p}\right)}\right)}{m^2}} \stackrel{(c)}{=} \O\left(\frac{1}{\sqrt n}\right),
    \label{eq:exp-decay-second-addend}
  \end{align}
  where (a) uses the exponential decay assumption. (b) uses that $n\ge m$ and that the logarithm is a monotonically increasing function. (c) follows from our choice of 
  \begin{align}
    m\gtrsim \sqrt{n}\sqrt{\log\left(1+\frac{c_1n}{c\trace\left(C_{\Q,\bar h_p}\right)}\right)\log(12n/\delta)\log(12/\delta)},
  \end{align}
  finishing the proof of the corollary.
\end{itemize}

\subsection{Proof of Theorem~\ref{thm:v-stat-consistency}}\label{sec:proof-v-stat-consistency}
    By the reverse triangle inequality, we obtain
    \begin{align}
        \left|S_p(\Q) - S_p\left( \hat \Q_n \right)\right| \le \norm{\mu_{h_p}(\Q) - \mu_{h_p}\left(\hat \Q_n\right)}{\H_{h_\P}} \hspace{-0.1cm}= \Bigg\|\frac1n\sum_{i=1}^{n}\underbrace{\left[h_p(\cdot, X_i) - \E_{X\sim\Q} h_p(\cdot, X)\right]}_{=:\eta_i}\Bigg\|_{\H_{h_p}},
    \end{align}
    which measures the concentration of i.i.d.\ centered random variables. To obtain the bound, we will use Bernstein's inequality (recalled in Theorem~\ref{thm:bernstein-hilbert}) by gaining control on the moments of $\norm{\eta_i}{\H_ {hp}}$ with Lemma~\ref{lemma:sub-exp-bernstein}.

    First, note that the $\norm{\eta_i}{\H_ {h_p}}$-s ($i\in[n]$) are sub-Gaussian as
    \begin{align}
        \norm{\norm{\eta_i}{\H_ {h_p}}}{\psi_2}&\stackrel{(a)}{=}\norm{\norm{\fm{X_i} - \E_{X\sim\Q} \fm X}{\H_{h_p}}}{\psi_2} \\
        &\stackrel{(b)}{\le} \norm{\norm{\fm{X_i}}{\H_{h_p}}  +\norm{\E_{X\sim\Q} \fm X}{\H_{h_p}}}{\psi_2} \\
        &\stackrel{(c)}{\le}  \norm{\norm{\fm{X_i}}{\H_{h_p}}  +\E_{X\sim\Q}\norm{ \fm X}{\H_{h_p}}}{\psi_2} \stackrel{(d)}{\lesssim} \norm{\norm{\fm{X_i}}{\H_{h_p}}}{\psi_2} < \infty.
    \end{align}
    We use the definition of $\eta_i$ in (a). (b) is implied by the triangle inequality and the monotonicity of the norm. (c) is by Jensen's inequality holding for Bochner integrals, and (d) follows from Lemma~\ref{lemma:orlicz-properties}(1); finiteness is due to the imposed assumption.

    Hence, $\norm{\eta_i}{\H_ {h_p}}$ is sub-exponential (Lemma~\ref{lemma:orlicz-properties}(3)), and, by Lemma~\ref{lemma:sub-exp-bernstein}, it holds for any $p\ge 2$ that
    \begin{align}
        \E_{X\sim\Q} \norm{\eta_i}{\H_ {h_p}}^p \le \frac12p!\sigma^2B^{p-2},
    \end{align}
    with $\sigma, B \lesssim \norm{\norm{\eta_i}{\H_ {h_p}}}{\psi_1} =: K$. Now,  applying  Theorem~\ref{thm:bernstein-hilbert} yields that, for any $\delta \in (0,1)$, it holds with probability at least $1-\delta$ that
    \begin{align}
        \norm{\frac 1n \sum_{i=1}^{n}\eta_i}{\H_{h_p}} \lesssim \frac{2K\log(2/\delta)}{n} + \sqrt{\frac{2K^2\log(2/\delta)}{n}},
    \end{align}
    which is the stated claim.

\section{AUXILIARY RESULTS}\label{sec:auxiliary-results}

This section collects our auxiliary results. Lemma~\ref{lemma:bound-projection} builds on \citet[Lemma~6]{rudi15less}, which assumes bounded feature maps, and on \citet[Lemma~5]{della21regularized}, which is stated in the context of leverage scores. The main technical challenge that we resolve lies in introducing and handling the centered covariance operator that allows us to make use of existing concentration results. %
Lemma~\ref{lemma:sub-exp-bernstein} states that a sub-exponential random variable satisfies Bernstein's conditions, and Lemma~\ref{lemma:sub-gauss-norm} is about the sub-Gaussianity of norms of Hilbert space-valued random variables. In Lemma~\ref{lemma:flip}, we show how tensor products interplay with linearly transformed vectors.  Lemma~\ref{lemma:max-of-sub-gauss} is about the maximum of real-valued sub-Gaussian random variables; it is a slightly altered restatement of \citet{canonne21tail}. In Lemma~\ref{lemma:prop-pos-operators} and Lemma~\ref{eq:cov-inequalities}, we collect inequalities of positive operators and of norms of covariance operators, respectively.

\begin{lemmaA}[Projected covariance operator bound] \label{lemma:bound-projection}
  Let Assumption 1 hold, and assume $0<\lambda\le\opnorm{C_{\Q,\bar h_p}}$. Then, for any $\delta \in (0,1)$, it holds that
    \begin{align}
        \left(\P^n \otimes \Lambda^m\right)\left(\opnorm{\left(I-P_{\H_{h_p,m}}\right)C_{\Q,\bar h_p,\lambda}^{1/2}}^2 \lesssim \lambda\right) \ge 1-\delta,
    \end{align}
    provided that $m\gtrsim \max\left\{ \frac{\trace\left(C_{\Q,\bar h_p}\right)}{\lambda},1  \right\}\log\left(4/\delta\right)$.

    \begin{proof}
      The proof proceeds in two steps: First, we show that $\opnorm{\left(I-P_{\H_{h_p,m}}\right)C_{\Q,\bar h_p,\lambda}^{1/2}}^2 \le \frac{\lambda}{1-\beta(\lambda)}$,
      when
          $\beta(\lambda) := \lambda_{\text{max}}\left(C_{\Q,\bar h_p,\lambda}^{-1/2}\left(C_{\Q,\bar h_p} - C_{\tilde \Q_m,\tilde h_p}\right)C_{\Q,\bar h_p,\lambda}^{-1/2}\right) <1$, where
          \begin{align}
          \tilde h_p(\cdot, \b x) &:= \fm {\b x} - \frac{1}{m} \sum_{i\in[m]} \fm{\tilde{\b x}_i} \quad (\b x \in \R^d),\\
              C_{\tilde \Q_m,\tilde h_p} &= \frac{1}{m}\sum_{i\in[m]} \tilde h_p\left(\cdot,\tilde{\b x}_i\right) \otimes  \tilde h_p\left(\cdot,\tilde{\b x}_i\right) \\
              &= \frac1m\sum_{i\in[m]} \fm{\tilde {\b x}_i} \otimes \fm{\tilde {\b x}_i} - \left(\frac1m\sum_{i\in[m]} \fm{\tilde {\b x}_i}\right) \otimes \left(\frac1m\sum_{i\in[m]} \fm{\tilde {\b x}_i}\right).
          \end{align}
          In the second step, we show that $\beta(\lambda) <1$ (with high probability) for $m$ large enough.

      \paragraph{Step 1.}

      Define the sampling operator $Z_m : \H_{h_p}\to \R^m$ by $f \mapsto \frac{1}{\sqrt m} \left(f(\tilde {\b x}_i)\right)_{i=1}^m$.
      Its adjoint $Z^*_m : \R^m \to \H_{h_p}$ (see \citet[Lemma~A.7(i)]{sterge22nystroem} is given by $\bm \alpha = \left(\alpha_i\right)_{i=1}^m \mapsto \frac{1}{\sqrt m}\sum_{i=1}^m\alpha_i\fm{\tilde{\b x}_i}$. Recall that $\H_{h_p,m}=\Span\left\{\fm{\tilde {\b x}_i}\mid i \in [m]\right\}$ and notice that $\operatorname{range}P_{\H_{h_p,m}} = \overline{\operatorname{range}Z^*_m}$. We obtain
      \begin{align} \label{eq:rudi-apply-prop3}
          \opnorm{\left(I-P_{\H_{h_p,m}}\right)C_{\Q,\bar h_p,\lambda}^{1/2}}^2 &\stackrel{(a)}{\le} \lambda\opnorm{\left(Z^*_mZ_m+\lambda I\right)^{-1/2}C_{\Q,\bar h_p,\lambda}^{1/2}}^2
          \stackrel{(b)}{=} \lambda\opnorm{ C_{\tilde \Q_m,h_p,\lambda}^{-1/2}C_{\Q,\bar h_p,\lambda}^{1/2}}^2 \\
          &\stackrel{(c)}{\le} \lambda \opnorm{ C_{\tilde \Q_m,\tilde h_p,\lambda}^{-1/2}C_{\Q,\bar h_p,\lambda}^{1/2}}^2
      \end{align}
      where (a) follows from \citet[Proposition~3]{rudi15less} with $X:=C_{\Q,\bar h_p,\lambda}^{1/2}$ therein. (b) is by \citet[Lemma~A.7(iv)]{sterge22nystroem}.
      Lemma~\ref{lemma:prop-pos-operators}(5) with $C:=C_{\tilde \Q_m,h_p,\lambda}^{-1/2}$, $D:=C_{\tilde \Q_m,\tilde h_p,\lambda}^{-1/2}$, and $X:= C_{\Q,\bar h_p,\lambda}^{1/2}$ yields (c), as we obtain $C^*C = C_{\tilde \Q_m,h_p,\lambda}^{-1}\preccurlyeq C_{\tilde \Q_m,\tilde h_p,\lambda}^{-1} = D^*D$; the positive definite relationship holding by the following chain of inequalities
      \begin{align}
          \MoveEqLeft C_{\tilde \Q_m,h_p,\lambda}^{-1}\preccurlyeq C_{\tilde \Q_m,\tilde h_p,\lambda}^{-1} \stackrel{\text{Lemma}~\ref{lemma:prop-pos-operators}(4)}{\iff} C_{\tilde \Q_m,h_p,\lambda}\succcurlyeq C_{\tilde \Q_m,\tilde h_p,\lambda} \stackrel{(d)}{\iff} C_{\tilde \Q_m,h_p}\succcurlyeq C_{\tilde \Q_m,\tilde h_p} \\
          &\stackrel{(e)}{\iff} 0 \preccurlyeq \mu_{h_p}\left(\tilde \Q_m\right) \otimes \mu_{h_p}\left(\tilde \Q_m\right),
      \end{align}
      which is true as the r.h.s.\ is a positive operator. In (d), we subtract $\lambda I$ from both sides. (e) follows from subtracting $C_{\tilde \Q_m,h_p}$ and by multiplying with $-1$.

      Applying the second inequality in the statement of \citet[Proposition~7]{rudi15less} to \eqref{eq:rudi-apply-prop3} (we specialize $A:= C_{\tilde \Q_m,\tilde h_p}$ and $B:=C_{\Q,\bar h_p}$ therein), we obtain
      \begin{align}
          \lambda\opnorm{ C_{\tilde \Q_m,\tilde h_p,\lambda}^{-1/2}C_{\Q,\bar h_p,\lambda}^{1/2}}^2 \le \frac{\lambda}{1-\beta(\lambda)}, \label{eq:prop3-7-2nd}
      \end{align}
      when $\beta(\lambda)<1$. The combination of \eqref{eq:rudi-apply-prop3} and \eqref{eq:prop3-7-2nd} yields the first stated claim.

      \paragraph{Step 2.} It remains to show that $\beta(\lambda) < 1$ holds with high probability. Let us introduce the shorthands $\tilde \mu_{h_p} = \mu_{h_p}\left(\tilde \Q_m\right) = \frac{1}{m}\sum_{i\in[m]}\fm{\tilde{\b x}_i}$ and $\mu_{h_p} = \mu_{h_p}(\Q)$. Notice that we have
      \begin{align} \label{eq:cov-identity}
          C_{\tilde \Q_m,\tilde h_p} = C_{\tilde \Q_m,\bar h_p} - \left[\tilde \mu_{h_p} - \mu_{h_p}\right] \otimes \left[\tilde \mu_{h_p} - \mu_{h_p}\right],
      \end{align}
      which is verified by using the linearity of tensor products and by using that 
      \begin{align}
        C_{\tilde \Q_m,\bar h_p} = \frac{1}{m}\sum_{i\in [m]} \fmc{\tilde{\b x}_i} \otimes \fmc{\tilde{\b x}_i}.
      \end{align}

      Instead of showing that $\beta(\lambda) < 1$, we will show that the following stronger requirement can be satisfied:
       \begin{align}
          \MoveEqLeft\beta(\lambda)
          \stackrel{(a)}{\le} \opnorm{C_{\Q,\bar h_p,\lambda}^{-1/2}\left(C_{\Q,\bar h_p} - C_{\tilde \Q_m,\tilde h_p}\right)C_{\Q,\bar h_p,\lambda}^{-1/2}} \\
          &\stackrel{(b)}{=} \opnorm{C_{\Q,\bar h_p,\lambda}^{-1/2}\left(C_{\Q,\bar h_p} - C_{\tilde \Q_m,\bar h_p} + \left[\tilde \mu_{h_p} - \mu_{h_p}\right] \otimes \left[\tilde \mu_{h_p} - \mu_{h_p}\right]\right)C_{\Q,\bar h_p,\lambda}^{-1/2}}\hspace*{-0.3cm} \\
          &\stackrel{(c)}{\le} \opnorm{C_{\Q,\bar h_p,\lambda}^{-1/2}\left(C_{\Q,\bar h_p} - C_{\tilde \Q_m,\bar h_p}\right)C_{\Q,\bar h_p,\lambda}^{-1/2}} +\\
          &\quad+ \opnorm{C_{\Q,\bar h_p,\lambda}^{-1/2}\left(\left[\tilde \mu_{h_p} - \mu_{h_p}\right] \otimes \left[\tilde \mu_{h_p} - \mu_{h_p}\right]\right)C_{\Q,\bar h_p,\lambda}^{-1/2}} \\
          &\stackrel{(d)}{=} \underbrace{\opnorm{C_{\Q,\bar h_p,\lambda}^{-1/2}\left(C_{\Q,\bar h_p} - C_{\tilde \Q_m,\bar h_p}\right)C_{\Q,\bar h_p,\lambda}^{-1/2}}}_{=:T_1} + \underbrace{\norm{C_{\Q,\bar h_p,\lambda}^{-1/2}\left(\tilde \mu_{h_p} - \mu_{h_p}\right)}{\H_{h_p}}^2}_{=:T_2} < 1.
      \end{align}
      In (a), we use that the spectral radius is bounded by the operator norm. (b) uses \eqref{eq:cov-identity} and (c) holds by the triangle inequality. Lemma~\ref{lemma:flip} and Lemma~\ref{lemma:norm-equiv} applied to the second term yield~(d).

      \begin{itemize}
          \item \tb{First term ($T_1$).} We will bring ourselves into the setting of  \citet[Theorem~9]{koltchinskii17concentration} (recalled in Theorem~\ref{thm:koltchinskii}). First, we condition on the Nyström selection and define the centered random variables $\eta_{i_j} = C_{\Q,\bar h_p,\lambda}^{-1/2}\left( \fm{\tilde{\b x}_{j}} - \E_{X\sim \Q}\fm{X} \right)$ ($=C_{\Q,\bar h_p,\lambda}^{-1/2}\bar h_p(\cdot,\tilde{\b x}_{j})$) ($j\in [m]$), which satisfy the sub-Gaussian assumption. Indeed, let $u\in \H_{h_p}$ be arbitrary, and $v= C_{\Q,\bar h_p,\lambda}^{-1/2}u \in \H_{h_p}$; the invertibility of $C_{\Q,\bar h_p,\lambda}$ guarantees the well-definedness of  $v$. With this notation, for any $j\in[m]$,
          \begin{align}
              \MoveEqLeft\norm{\ip{\eta_{i_j},u}{\H_{h_p}}}{\psi_2}
              \stackrel{(a)}{=} \norm{\ip{C_{\Q,\bar h_p,\lambda}^{-1/2}\bar h_p(\cdot,\tilde{\b x}_j),u}{\H_{h_p}}}{\psi_2}
              \stackrel{(b)}{=} \norm{\ip{\bar h_p(\cdot,\tilde{\b x}_{j}),C_{\Q,\bar h_p,\lambda}^{-1/2}u}{\H_{h_p}}}{\psi_2} \\
              &\stackrel{(c)}{=} \norm{\ip{\bar h_p(\cdot,\tilde{\b x}_{j}),v}{\H_{h_p}}}{\psi_2}
              \hspace{-0.2cm}\stackrel{(d)}{\lesssim} \underbrace{\norm{\ip{\bar h_p(\cdot,\tilde{\b x}_{j}),v}{\H_{h_p}}}{L_2(\Q)}}_{(\dagger)}
              \hspace{-0.2cm}\stackrel{(e)}{=} \norm{\ip{\bar h_p(\cdot,\tilde{\b x}_{j}),C_{\Q,\bar h_p,\lambda}^{-1/2}u}{\H_{h_p}}}{L_2(\Q)} \\
              &\stackrel{(f)}{=} \norm{\ip{C_{\Q,\bar h_p,\lambda}^{-1/2}\bar h_p(\cdot,\tilde{\b x}_{i_j}),u}{\H_{h_p}}}{L_2(\Q)}
              \stackrel{(g)}{=} \norm{\ip{\eta_{i_j},u}{\H_{h_p}}}{L_2(\Q)} < \infty.
          \end{align}
          (a) is the definition of the $\eta_{i_j}$-s, (b) uses the self-adjointness of $C_{\Q,\bar h_p,\lambda}$, and (c) follows from the definition of $v$. The sub-Gaussian assumption implies (d), (e) again follows from the definition of $v$, and (f) is implied by the self-adjointness of $C_{\Q,\bar h_p,\lambda}$. Inserting the definition of $\eta_{i_j}$ in (g) proves their sub-Gaussianity  by using that $(\dagger) <\infty$ according to Assumption~\ref{ass:sub-gaussian} and as the derivation afterwards only involved equalities.

          Let $A=C_{\Q,\bar h_p,\lambda}^{-1/2}C_{\Q,\bar h_p}C_{\Q,\bar h_p,\lambda}^{-1/2}$. Theorem~\ref{thm:koltchinskii} yields that, conditioned on the Nyström selection, it holds with probability at least $1-\delta/2$ that
          \begin{align}
              \Big\|\underbrace{C_{\Q,\bar h_p,\lambda}^{-1/2}\left(C_{\Q,\bar h_p} - C_{\tilde \Q_m,\bar h_p}\right)C_{\Q,\bar h_p,\lambda}^{-1/2}}_{= \frac{1}{m}\sum_{j=1}^m\eta_{i_j}\otimes\eta_{i_j} - \E \left[\eta_{i_j}\otimes \eta_{i_j}\right]}\Big\|_{\text{op}} \lesssim \opnorm{A}\max\left(\sqrt{\frac{r\left(A\right)}{m}},\sqrt{\frac{\log(2/\delta)}{m}}\right),
          \end{align}
          provided that $m\ge \max\left\{r\left(A\right),\log(2/\delta)\right\}$, with $r\left(A\right) = \frac{\trace \left( A \right)}{\opnorm{A}}$. Using Lemma~\ref{lemma:prop-pos-operators}(2), $A  \preccurlyeq I$, hence $\opnorm{A} \le 1$. Moreover by Lemma~\ref{eq:cov-inequalities}(3), $r(A) \le \frac{2\trace\left(C_{\Q,\bar h_p}\right)}{\lambda}$, which implies that, with the same probability,
          \begin{align}
              \opnorm{C_{\Q,\bar h_p,\lambda}^{-1/2}\left(C_{\Q,\bar h_p} - C_{\tilde \Q_m,\bar h_p}\right)C_{\Q,\bar h_p,\lambda}^{-1/2}} \lesssim \max\left(\sqrt{\frac{\trace\left(C_{\Q,\bar h_p}\right)}{\lambda m}},\sqrt{\frac{\log(2/\delta)}{m}}\right),
          \end{align}
          holds when $m\ge \max\left\{\frac{2\trace\left(C_{\Q,\bar h_p}\right)}{\lambda},\log(2/\delta)\right\}$. Therefore, one can take 
          \begin{align}
            m\gtrsim \max\left\{\frac{\trace\left(C_{\Q,\bar h_p}\right)}{\lambda},\log(2/\delta)\right\}
          \end{align}
          to get $\opnorm{C_{\Q,\bar h_p,\lambda}^{-1/2}\left(C_{\Q,\bar h_p} - C_{\tilde \Q_m,\bar h_p}\right)C_{\Q,\bar h_p,\lambda}^{-1/2}} < \frac12$ holding with probability at least $1-\delta/2$.

          \item \tb{Second term ($T_2$).} We condition again on the Nyström selection, let $\eta_{i_j} = C_{\Q,\bar h_p,\lambda}^{-1/2}\bar h_p\left(\cdot, \b x_{i_j}\right)$ for $j\in[m]$, and observe that $\frac1m \sum_{j\in[m]}\eta_{i_j} = C_{\Q,\bar h_p,\lambda}^{-1/2}\left( \tilde \mu_{h_p}- \mu_{h_p}\right)$. The $\eta_{i_j}$-s are centered, and, by Lemma~\ref{lemma:sub-gauss-norm}, it holds for any $j\in[m]$ that
          \begin{align}
              \norm{\norm{\eta_{i_j}}{\H_{h_p}}}{\psi_2}^2 \lesssim \trace \left( C_{\Q,\bar h_p,\lambda}^{-1}C_{\Q,\bar h_p} \right),
          \end{align}
          that is, the $\norm{\eta_{i_j}}{\H_{h_p}}$-s are sub-Gaussian. Hence, by Lemma~\ref{lemma:orlicz-properties}(3), they are sub-exponential, and, by Lemma~\ref{lemma:sub-exp-bernstein}, they satisfy the Bernstein condition
          with $\sigma, B \lesssim \sqrt{\trace\left( C_{\Q,\bar h_p,\lambda}^{-1}C_{\Q,\bar h_p} \right)}$. Therefore, application of Theorem~\ref{thm:bernstein-hilbert} yields that, conditioned on the Nyström choice, it holds with probability at least $1-\delta/2$ that
          \begin{align}
              \norm{\frac1m\sum_{j=1}^{m}\eta_{i_j}}{\H_{h_p}}
              &\lesssim \frac{\sqrt{\trace\left( C_{\Q,\bar h_p,\lambda}^{-1}C_{\Q,\bar h_p} \right)}\log(4/\delta)}{m} + \sqrt{\frac{\trace\left( C_{\Q,\bar h_p,\lambda}^{-1}C_{\Q,\bar h_p} \right)\log(4/\delta)}{m}} \\
              &\stackrel{(a)}{\lesssim} \sqrt{\frac{\trace\left( C_{\Q,\bar h_p,\lambda}^{-1}C_{\Q,\bar h_p} \right)\log(4/\delta)}{m}}
              \stackrel{(b)}{\le} \sqrt{\frac{\trace\left(C_{\Q,\bar h_p} \right)\log(4/\delta)}{\lambda m}}
          \end{align}
          where in (a), we assume that $m\ge \log(4/\delta)$ and notice that this condition implies that the first term is smaller than the second term. Lemma~\ref{eq:cov-inequalities}(1) yields (b). The obtained bound means that choosing $m\gtrsim \max\left\{ \frac{\trace\left(C_{\Q,\bar h_p}\right)}{\lambda},1  \right\}\log(4/\delta)$ guarantees that $\norm{\frac1m\sum_{j=1}^{m}\eta_{i_j}}{\H_{h_p}}^2 < \frac{1}{2}$ holds with probability at least $1-\delta/2$.

      \end{itemize}

      As a final step, we observe that $\log(2/\delta) < \log(4/\delta)$ and $\log(4/\delta) > 1$, which, by union bound, shows that, for $m\gtrsim \max\left\{ \frac{\trace\left(C_{\Q,\bar h_p}\right)}{\lambda},1  \right\}\log(4/\delta)$, it holds with probability at least $1-\delta$ that $\beta(\lambda) < 1$. We lift the conditioning by integrating over all Nyström selections.
      \qedhere
  \end{proof}
\end{lemmaA}

\begin{lemmaA}[Sub-exponential satisfies Bernstein conditions]\label{lemma:sub-exp-bernstein}
Let $Y$ be a real-valued random variable which is sub-exponential, i.e.\ $\norm{Y}{\psi_1} < \infty$. Let
  $ \sigma := \sqrt 2 \norm{Y}{\psi_1}$, $ B:= \norm{Y}{\psi_1} >0$. Then the Bernstein condition
  \begin{align}
    \E|Y|^p \le \frac{1}{2}p!\sigma^2B^{p-2} < \infty \label{eq:Bernstein}
    \end{align}
  holds for any $p\ge 2$.
  \begin{proof} For any $p\ge 2$, we have
    \begin{align}
      \E|Y|^p = p! B^p \E \frac{|Y|^p}{B^p p!} \stackrel{(a)}{<} p! B^p \underbrace{\left[\E \exp\left(\frac{|Y|}{B}\right) -1\right]}_{\stackrel{(b)}{\le}1} = \frac{1}{2}p!B^{p-2}\left(\sqrt 2 B\right)^2,
    \end{align}
    where in (a) we use that $\frac{x^n}{n!} < e^x-1$ holds for all $n,x > 0$, and (b) follows from the definition of the sub-exponential Orlicz norm.
  \end{proof}

\end{lemmaA}

The next lemma shows that  $\fmc X$ and the ``whitened'' random variable $C_{\Q,\bar h_p,\lambda}^{-1/2}\fmc X$ enjoy sub-Gaussian properties in terms of their respective $\H_{h_p}$ norms.

\begin{lemmaA}[Sub-Gaussianity of norm of Hilbert space-valued random variables] \label{lemma:sub-gauss-norm}
   Let $\H$ be a separable Hilbert space, $Y \sim \Q \in \mathcal M_1^+(\H)$, and $A \in \mathcal L(\H)$ invertible, and positive. Assume that $Y$ is sub-Gaussian, in other words $\norm{\ip{Y,u}{\H}}{\psi_2} \lesssim \norm{\ip{Y,u}{\H}}{L_2(\Q)}$ holds for all $u\in\H$. Then
  \begin{align}
    \norm{\norm{A^{1/2}Y}{\H}}{\psi_2}^2 \lesssim \trace \left(A\E_{Y\sim\Q} \left(Y\otimes Y\right)\right).
  \end{align}
  Specifically, with Assumption~\ref{ass:sub-gaussian}, choosing $A:= I$ and $Y:=\fmc X$, and $A:=C^{-1}_{\Q,\bar h_p,\lambda}$ ($\lambda > 0$) and $Y:=\fmc X$, respectively, it holds that
  \begin{align}
    \norm{\norm{\fmc X}{\H_{h_p}}}{\psi_2} < \infty, && \text{and} && \norm{\norm{C_{\Q,\bar h_p,\lambda}^{-1/2}\fmc X}{\H_{h_p}}}{\psi_2}^2 \lesssim \trace \left( C_{\Q,\bar h_p,\lambda}^{-1}C_{\Q,\bar h_p} \right) < \infty,
  \end{align}
  that is, both  $\norm{\fmc X}{\H_{h_p}}$ and $\norm{C_{\Q,\bar h_p,\lambda}^{-1/2}\fmc X}{\H_{h_p}}$ are sub-Gaussian.
\end{lemmaA}
\begin{proof}
  Let $ \left( e_i \right)_{i\in I}$ be a countable ONB of the separable $\H$. We obtain
  \begin{align}
    \MoveEqLeft\norm{\norm{A^{1/2}Y}{\H}}{\psi_2}^2
    \stackrel{(a)}{=} \norm{\norm{A^{1/2}Y}{\H}^2}{\psi_1}
    \stackrel{(b)}{=} \norm{\sum_{i\in I}\ip{A^{1/2}Y,e_i}{\H}^2}{\psi_1}
    \stackrel{(c)}{\le} \sum_{i\in I}\norm{\ip{A^{1/2}Y,e_i}{\H}^2}{\psi_1} \\
    &\stackrel{(d)}{=}  \sum_{i\in I}\norm{\ip{A^{1/2}Y,e_i}{\H}}{\psi_2}^2
    \stackrel{(e)}{\lesssim} \sum_{i\in I}\norm{\ip{A^{1/2}Y,e_i}{\H}}{L_2(\Q)}^2
    \stackrel{(f)}{=} \sum_{i\in I}\E_{Y\sim\Q}\ip{A^{1/2}Y,e_i}{\H}^2 \\
    &\stackrel{(g)}{=} \sum_{i\in I}\E_{Y\sim\Q}\ip{\left(A^{1/2}Y\right)\otimes \left(A^{1/2}Y\right),e_i\otimes e_i}{\H\otimes \H} \\
    &\stackrel{(h)}{=} \sum_{i\in I}\E_{Y\sim\Q}\ip{A^{1/2}\left(Y\otimes Y\right)A^{1/2},e_i\otimes e_i}{\H\otimes \H} \\
    &\stackrel{(i)}{=} \sum_{i\in I}\ip{A^{1/2}\E_{Y\sim\Q}\left(Y\otimes Y\right)A^{1/2},e_i\otimes e_i}{\H\otimes \H}
    \stackrel{(j)}{=} \sum_{i\in I}\ip{A^{1/2}\E_{Y\sim\Q}\left(Y\otimes Y\right)A^{1/2}e_i, e_i}{\H} \\
    &\stackrel{(k)}{=} \trace \left( A^{1/2}\E_{Y\sim\Q}\left(Y\otimes Y\right)A^{1/2} \right)
    \stackrel{(l)}{=} \trace \left( A\E_{Y\sim\Q}\left(Y\otimes Y\right) \right).
\end{align}
The details are as follows.
(a) uses Lemma~\ref{lemma:orlicz-properties}(4),
Parseval's identity yields (b),
and the triangle inequality implies (c).
(d) holds by Lemma~\ref{lemma:orlicz-properties}(4).
For (e), let $u_i=A^{1/2}e_i$ and observe that
\begin{align}
  \norm{\ip{A^{1/2}Y,e_i}{\H}}{\psi_2}^2 \stackrel{(m)}{=} \norm{\ip{Y,u_i}{\H}}{\psi_2}^2 \stackrel{(n)}{\lesssim} \norm{\ip{Y,u_i}{\H}}{L_2(\Q)}^2 \stackrel{(o)}{=} \norm{\ip{A^{1/2}Y,e_i}{\H}}{L_2(\Q)}^2,
\end{align}
where (m) uses the self-adjointness of $A^{1/2}$ (implied by the positivity of $A$), (n) follows from the sub-Gaussian assumption on $Y$ holding for arbitrary $u_i\in \H$, and (o), again, uses the self-adjointness of $A$.
(f) is the definition of the $L_2(\Q)$-norm, (g) holds by the definition of the tensor product, and Lemma~\ref{lemma:flip} yields (h).
(i) integral and bounded linear operators are swapped by  \citet[(A.32)]{steinwart08support}, (j) is a property of Hilbert-Schmidt operators, and (k) uses the definition of the trace of a linear operator w.r.t.\ an ONB. The cyclic invariance property of the trace yields (l) and concludes the proof of the first statement.

With $A:= I$ and $Y:=\fmc X$, we have $\norm{\norm{\fmc X}{\H_{h_p}}}{\psi_2} \hspace{-0.1cm}\lesssim \trace \left( \E \left(\fmc X \otimes \fmc X\right) \right) = \trace (C_{\Q,\bar h_p}) < \infty$, which is the second statement.
The last part follows from considering $A:=C^{-1}_{\Q,\bar h_p,\lambda}$ and $Y:=\fmc X$; the invertibility of $C_{\Q,\bar h_p,\lambda}$ guarantees the well-definedness of the $u_i$-s ($i\in I$).
\end{proof}

The following lemma is a natural generalization of the property $(\b{Ca})(\b{Db})\T = \b{C}\left(\b{ab}\T\right)\b{D}\T$, where $\b C,\b D\in \R^{d\times d}$ and $\b a,\b b\in \R^d$.
\begin{lemmaA}[Tensor product of linearly transformed vectors] \label{lemma:flip} Let $\H$ be a Hilbert space and $C,D \in \L(\H)$. Then for any $a,b\in \H$,  $(Ca) \otimes (Db) = C (a\otimes b) D^*$. Specifically, when $D$ is self-adjoint, it holds that $(Ca) \otimes (Db) = C (a\otimes b) D$.
\begin{proof}
Let $h\in \H$ be arbitrary and fixed. Then,
\begin{align*}
[(Ca) \otimes (Db)](h) & \stackrel{(a)}{=}  Ca \langle Db,h\rangle_{\H},\\
[C (a\otimes b) D^*](h) & = C(a\otimes b) (D^* h) \stackrel{(b)}{=} Ca \langle b, D^*h\rangle_{\H} \stackrel{(c)}{=} Ca \langle Db,h\rangle_{\H}.
\end{align*}
In (a) and (b), we used the definition of $\otimes$, (c) follows from the definition of the adjoint and by the property $(D^*)^*=D$. The  shown equality of $[(Ca) \otimes (Db)](h) = [C (a\otimes b) D^*](h)$ for any $h\in \H$ proves the claimed statement.
\end{proof}
\end{lemmaA}

\begin{lemmaA}[Maximum of sub-Gaussian random variables]\label{lemma:max-of-sub-gauss}
    Let $\left(X_i\right)_{i=1}^n \stackrel{\text{i.i.d.}}{\sim} \P$ be real-valued sub-Gaussian random variables. Then $\P\left(\max_{i\in[n]}|X_i| \lesssim \sqrt{\norm{X_1}{\psi_2}^2\log(2n/\delta)}\right) \ge 1-\delta$ holds for any $\delta\in(0,1)$.
    \begin{proof}
        Let $c>0$ be an absolute constant. As $X_1$ is sub-Gaussian, by \citet[Proposition~2.5.2]{vershynin18highdimensional}, there exists $K_1\le c\norm{X_1}{\psi_2}$ such that $\P(|X_1|\ge t) \le 2e^{-\frac{t^2}{K_1^2}}$ for all $t\ge 0$. Let $u= \sqrt{K_1^2(\log(2n) + t)}$. Then
        \begin{align}
            \P\left(\max_{i\in[n]}|X_i|\ge u\right) \stackrel{(a)}{\le} \sum_{i=1}^n \P\left(|X_i|
            \ge u\right) \stackrel{(b)}{\le} 2ne^{-\frac{u^2}{K_1^2}} \stackrel{(c)}{=} e^{-t},
        \end{align}
        where (a) uses that the probability of a maximum exceeding a value is less than the sum of the probability of its arguments exceeding the value, (b) uses the mentioned property of sub-Gaussian random variables, and (c) is our definition of $u$. Solving for $\delta := e^{-t}$ gives $t=\log(1/\delta)$, and considering the complement yields $\P\left(\max_{i\in[n]}|X_i|\le \sqrt{K_1^2\log(2n/\delta)}\right) \ge 1-\delta$. Using that $K_1 \le c\norm{X_1}{\psi_2}$ concludes the proof.
    \end{proof}
\end{lemmaA}

The following result shows that positive operators share some well-known properties of positive (semi-)definite matrices; we refer to \citet{bhatia07positive} for the related matrix cases.

\begin{lemmaA}[Properties of positive operators] \label{lemma:prop-pos-operators}
    Let $\H$ be a Hilbert space and assume $A,B \in \mathcal L\left(\H\right)$ are positive and invertible. Then, the following hold.
    \begin{enumerate}
        \item If $A \preccurlyeq B$, then $X^*AX \preccurlyeq X^*BX$ for any $X \in \mathcal L\left(\H\right)$.
        \item If $A \preccurlyeq B$, then $B^{-1/2}AB^{-1/2} \preccurlyeq I$.
        \item If $B \preccurlyeq I$, then $B^{-1} \succcurlyeq I$.
        \item If $A \preccurlyeq B$, then $A^{-1} \succcurlyeq B^{-1}$.
        \item If $C^*C \preccurlyeq D^*D$, then $\opnorm{CX} \le \opnorm{DX}$ for any $C,D,X \in \mathcal L\left(\H\right)$.
    \end{enumerate}
\end{lemmaA}
\begin{proof}~
    \begin{enumerate}
        \item For any $x \in \H$, one has $\ip{x,X^*AXx}{\H} = \ip{Xx,AXx}{\H} \stackrel{(\dagger)}{\le} \ip{Xx,BXx}{\H} = \ip{x,X^*BXx}{\H}$; $(\dagger)$ follows from $A \preccurlyeq B$ applied to $Xx$.
        \item We apply (1.) with $X=B^{-1/2}$.
        \item We have $B^{-1} = B^{-1/2}IB^{-1/2} \succcurlyeq B^{-1/2}BB^{-1/2} = I$, where we used (1.) in the second step.
        \item By (2.), it holds that $B^{-1/2}AB^{-1/2} \preccurlyeq I$, from which (3.) implies that $B^{1/2}A^{-1}B^{1/2} \succcurlyeq I$. Now apply (1.) with $X=B^{-1/2}$ to obtain the stated result.
        \item The $C^*$-property, the definition of the adjoint and that of the operator norm yield
        \begin{align}
            \MoveEqLeft\opnorm{CX}^2 = \opnorm{X^*C^*CX} = \sup_{\norm{x}{\H}=1}\ip{x,X^*C^*CXx}{\H} = \sup_{\norm{x}{\H}=1}\ip{Xx,C^*CXx}{\H} \\
            &\hspace{-0.1cm}\le \sup_{\norm{x}{\H}=1}\ip{Xx,D^*DXx}{\H} = \sup_{\norm{x}{\H}=1}\ip{x,X^*D^*DXx}{\H} = \opnorm{X^*D^*DX} = \opnorm{DX}^2,
        \end{align}
        which, after taking the positive square root, proves the claim.  \qedhere

    \end{enumerate}
\end{proof}

The following lemma collects some inequalities for the trace and operator norms of covariance operators. Many of these are known and frequently employed without proof; we provide proofs here for completeness.

\begin{lemmaA}[Covariance operator inequalities]\label{eq:cov-inequalities}
    Let $\H$ be a separable Hilbert space, $X\sim \Q \in \mathcal M_1^+(\H)$, $C_{\Q} = \E \left[X\otimes X\right]$, $C_{\Q,\lambda} = C_{\Q} + \lambda I$, and let $r(\cdot) = \frac{\trace\left(\cdot\right)}{\opnorm{\cdot}}$ be defined on trace-class operators. Assume that $0<\lambda \le \opnorm{C_{\Q}}$. Then, the following hold.
    \begin{enumerate}
        \item $\frac{1}{2}r\left(C_{\Q}\right) \le \trace \left( C_{\Q,\lambda}^{-1}C_{\Q} \right) \le \frac{\trace\left(C_{\Q}\right)}{\lambda}$,
        \item $\frac 12 \le \opnorm{C_{\Q,\lambda}^{-1/2}C_{\Q}C_{\Q,\lambda}^{-1/2}}< 1$, and
        \item $r\left(C_{\Q,\lambda}^{-1/2}C_{\Q}C_{\Q,\lambda}^{-1/2}\right) \le \frac{2\trace\left(C_{\Q}\right)}{\lambda}$.
    \end{enumerate}
\end{lemmaA}
\begin{proof} Let $\left(\lambda_i\right)_{i\in I}$ denote the eigenvalues of $C_\Q$, with $\lambda_1 \ge \lambda _2 \ge \cdots \ge 0$.
    \begin{enumerate}
        \item The first inequality follows from $\trace \left( C_{\Q,\lambda}^{-1}C_{\Q} \right) = \sum_{i\in I} \frac{\lambda_i}{\lambda_i + \lambda} \ge \sum_{i\in I}\frac{\lambda_i}{2\opnorm{C_{\Q}}} = \frac{\trace\left(C_\Q\right)}{2\opnorm{C_\Q}}$. The second one is footnote~\ref{fn:cov-ineq}.
        \item For the first inequality, observe that $\opnorm{C_{\Q,\lambda}^{-1/2}C_{\Q}C_{\Q,\lambda}^{-1/2}} = \frac{\lambda_1}{\lambda_1 + \lambda} \stackrel{(\dagger)}{\ge} \frac{1}{2}$, where $(\dagger)\Leftrightarrow 2\lambda_1 \ge \lambda_1 + \lambda \Leftrightarrow {(\opnorm{C_{\Q}}=)}\lambda_1 \ge \lambda$, which holds by assumption. The second one is implied as $\frac{\lambda_1}{\lambda_1 + \lambda} \stackrel{(\dagger)}{<} 1$, where $(\dagger) \Leftrightarrow \lambda_1 < \lambda_1 + \lambda \Leftrightarrow 0 < \lambda$; this condition was again assumed.
        \item We upper bound the numerator of $r(C_{\Q,\lambda}^{-1/2}C_{\Q}C_{\Q,\lambda}^{-1/2})$ by (1.) after, using the cyclic invariance of the trace, we rewrite it as $\trace \left(C_{\Q,\lambda}^{-1/2}C_{\Q}C_{\Q,\lambda}^{-1/2}\right) =\trace \left( C_{\Q,\lambda}^{-1}C_{\Q} \right)$, and lower bound the denominator by (2.).\qedhere
    \end{enumerate}
\end{proof}

\section{EXTERNAL STATEMENTS}\label{sec:external-statements}

This section collects the external statements that we use. Lemma~\ref{lemma:norm-equiv} gives equivalent norms for $f\otimes f$. We collect properties of Orlicz norms in Lemma~\ref{lemma:orlicz-properties}. Theorem~\ref{thm:koltchinskii} is about the concentration of the empirical covariance, and Theorem~\ref{thm:bernstein-hilbert} recalls Bernstein's inequality for separable Hilbert spaces. Theorem~\ref{thm:bernstein-bounded} is a concentration result for bounded random variables in a separable Hilbert space.

\begin{lemmaA}[Lemma B.8; \citealt{sriperumbudur22approximate}] \label{lemma:norm-equiv} Define $B = f \otimes f$, where $f\in\H$ and $\H$ is a separable Hilbert space. Then $\opnorm{B}=\hsnorm{B}{\H} = \trace B = \norm{f}{\H}^2$.
\end{lemmaA}

We refer to the following sources for the items in Lemma~\ref{lemma:orlicz-properties}. Item 1 is \citet[Lemma~2.6.8]{vershynin18highdimensional}, Item 2 is \citet[Exercise~2.7.10]{vershynin18highdimensional}, Item 3 recalls \citet[p.~95]{vaart96weak}, and Item 4 is \citet[Lemma~2.7.6]{vershynin18highdimensional}.

\begin{lemmaA}[Collection of Orlicz properties]\label{lemma:orlicz-properties} Let $X$ be a real-valued random variable.
  \begin{enumerate}
    \item If $X$ is sub-Gaussian, then $X-\E X$ is also sub-Gaussian, and
    \begin{align}
      \norm{X-\E X}{\psi_2} \le \norm{X}{\psi_2} + \norm{\E X}{\psi_2} \lesssim \norm{X}{\psi_2}.
    \end{align}
    \item  If $X$ is sub-exponential, then $X-\E X$ is also sub-exponential, and satisfies
    \begin{align}
      \norm{X-\E X}{\psi_1} \le \norm{X}{\psi_1} + \norm{\E X}{\psi_1} \lesssim \norm{X}{\psi_1}.
    \end{align}
    \item If $X$ is sub-Gaussian, it is sub-exponential. Specifically, it holds that $\norm{X}{\psi_1} \le \sqrt{\log 2}\norm{X}{\psi_2}$.
    \item $X$ is sub-Gaussian if and only if $X^2$ is sub-exponential. Moreover,
    \begin{align}
      \norm{X^2}{\psi_1} = \norm{X}{\psi_2}^2.
    \end{align}
  \end{enumerate}
\end{lemmaA}

\begin{theoremA}[Theorem 9; \citealt{koltchinskii17concentration}]\label{thm:koltchinskii}
  Let $X, X_1,\ldots, X_n$ be i.i.d.\ square integrable centered random vectors in a Hilbert space $\H$ with covariance operator $C$. Let the empirical covariance operator be $\hat C_n = \frac1n\sum_{i=1}^nX_i\otimes X_i$. If $X$ is sub-Gaussian, then there exists a constant $c>0$ such that, for all $\delta\in(0,1)$, with probability at least $1-\delta$,
  \begin{align}
    \opnorm{\hat C_n- C}\le c\opnorm{C}\max\left(\sqrt{\frac{r(C)}{n}},\;\sqrt{\frac{\log(1/\delta)}{n}}\right),
  \end{align}
  provided that $n\ge \max\{r(C),\log (1/\delta)\}$, where $r(C) := \frac{\trace(C)}{\opnorm{C}}$.
\end{theoremA}

The following theorem by \citet{yurinsky95sums} is quoted from \citet{sriperumbudur22approximate}.

\begin{theoremA}[Bernstein bound for separable Hilbert spaces; Theorem 3.3.4; \citealt{yurinsky95sums}] \label{thm:bernstein-hilbert}
  Let $\left( \Omega,\mathcal A,\P \right)$ be a probability space, $\H$ a separable Hilbert space, $B> 0$, $\sigma >0$, and $\eta_{1},\ldots,\eta_{n} : \Omega\to \H$ centered i.i.d.\ random variables that satisfy
  \begin{align*}
    \E\norm{\eta_{1}}{\H}^{p} \le \frac12p!\sigma^2B^{p-2}
  \end{align*}
  for all $p\ge 2$. Then, for any $\delta \in (0,1)$ it holds with probability at least $1-\delta$ that
  \begin{align*}
    \norm{\frac1n\sum_{i=1}^{n}\eta_{i}}{\H} \le \frac{2B\log(2/\delta)}{n}+\sqrt{\frac{2\sigma^{2}\log(2/\delta)}{n}}.
  \end{align*}
\end{theoremA}

\begin{theoremA}[Concentration in separable Hilbert spaces; Lemma~E.1; \citealt{chatalic22nystrom}]
  \label{thm:bernstein-bounded}
  Let $X_1,\dots,X_n$ be i.i.d.\ random variables with zero mean in a separable Hilbert space $\left(\H,\norm{\cdot}{\H}\right)$ such that $\max_{i\in[n]}\norm{X_i}{\H} \le b$ almost surely, for some  $b > 0$. Then for any $\delta \in (0,1)$, it holds with probability at least $1-\delta$ that
  \begin{align*}
    \norm{\frac 1n \sum_{i=1}^nX_i}{\H}\le b\frac{\sqrt{2\log (2/\delta)}}{\sqrt n}.
  \end{align*}
\end{theoremA}

\section{ADDITIONAL EXPERIMENTS}\label{sec:additional-results}

In this section, we collect additional numerical results. Section~\ref{sec:runtime-vs-power} discusses the trade-off between power and runtime of the tested approaches. Section~\ref{sec:impact-size-nystroem-sample} shows the impact of the size of the Nyström sample.

\subsection{Runtime vs.\ Power}\label{sec:runtime-vs-power}

Based on the experimental setup in Section~\ref{sec:experiments}, we performed an additional set of experiments to contrast runtime and power. We repeated each setup for $100$ rounds to obtain the given power and average runtime. The quadratic-time approaches are considered as baseline.

\begin{figure}[ht]
\centering
\includegraphics[width=.9\textwidth]{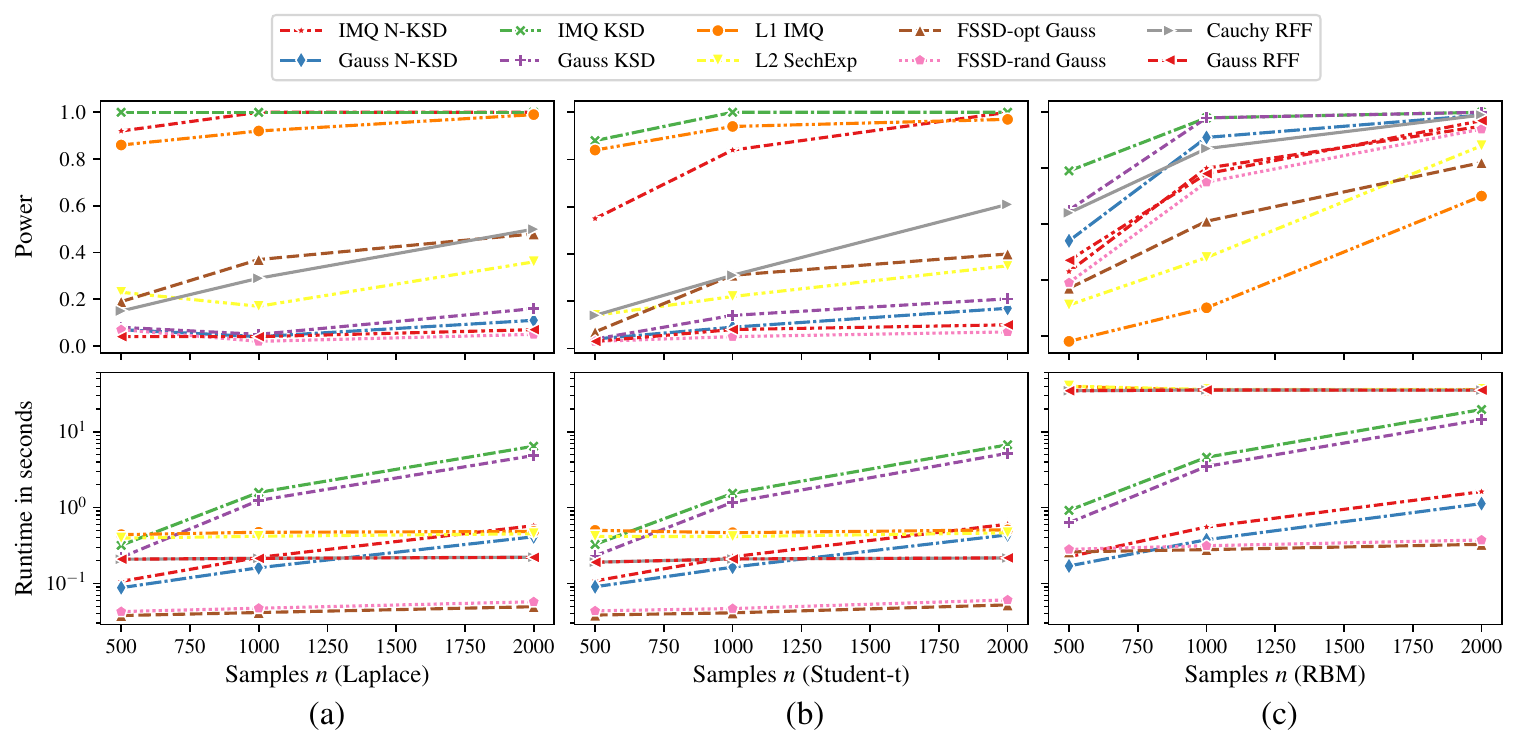}
\caption{Runtime and power trade-off of the tested approximations.}
\label{fig:runtime-vs-power}
\end{figure}

\paragraph{Laplace vs.\ standard normal.} We fix $d=15$, $m=4\sqrt n$, and vary $n\in\{500,1000,2000\}$. The remaining parameters match the ones stated in Section~\ref{sec:experiments}.

Figure~\ref{fig:runtime-vs-power}(a) summarizes our results regarding power and runtime. The results show that the proposed IMQ N-KSD approach has the highest power of all approximations across all tested $n$. Second best is L1 IMQ. W.r.t.\ runtime, the proposed method is faster than L1 IMQ for $n\in\{500,1000\}$. For $n=2000$, IMQ N-KSD has a similar runtime but still features better power. The FSSD approaches are the fastest but do not have a high power in this experiment.

\paragraph{Student-t vs.\ standard normal.} Again, we fix $d=15$, set $m=4\sqrt n$, and vary $n\in\{500,1000,2000\}$. The other parameters are the sames as the ones stated in Section~\ref{sec:experiments}.

Figure~\ref{fig:runtime-vs-power}(b) shows that L1 IMQ achieves higher power than the proposed IMQ N-KSD for $n\in\{500,1000\}$ but at the price of a larger runtime. For $n=2000$, the performance of IMQ N-KSD is slightly better than that of L1 IMQ while both approaches have a similar runtime. The remaining approaches perform worse in terms of power.

\paragraph{Restricted Boltzmann machine (RBM).} For the RBM experiment, we set $\sigma=0.02$, $m=4\sqrt n$, and select $n\in\{500,1000,2000\}$; all other parameters match the ones detailed in Section~\ref{sec:experiments}.

We summarize the results in Figure~\ref{fig:runtime-vs-power}(c). While both random feature Stein discrepancies (L1 IMQ, L2 SechExp) scale linearly in $n$, the higher dimensionality and difficulty of this problem result in a runtime that is orders of magnitude larger than that of all other approximations; the same observation w.r.t.\ runtime applies to the RFF approaches. We also observe that the runtimes of the related FSSD approaches increase compared to their runtime results in the Laplace and Student-t experiments.

Regarding power, the proposed Gauss N-KSD achieves the best result of all approximations from $n\ge 1000$ while being among the fastest methods. While, for $n\in\{1000,2000\}$, it is a bit slower than the FSSD approaches, the proposed method achieves higher power across all choices of $n$.

\tb{Summary.} Figure 2(a)--(b) shows that some existing methods, e.g., L1 IMQ, perform similarly to N-KSD in terms of power achieved but come with a larger runtime for smaller sample sizes. Figure 2(c) highlights that some competitors (L1 IMQ, L2 SechExp, Cauchy RFF, Gauss RFF) require a larger runtime than the baseline approaches (IMQ KSD, Gauss KSD) for samples of size less than 2000. Here, our method is one to two orders of magnitude faster while achieving the same or, in some cases, larger power.

These results show that the proposed N-KSD has a very good runtime/power trade-off.

\subsection{Impact of the Size of the Nyström Sample}\label{sec:impact-size-nystroem-sample}

\begin{figure}[ht]
    \centering
    \includegraphics[width=.9\textwidth]{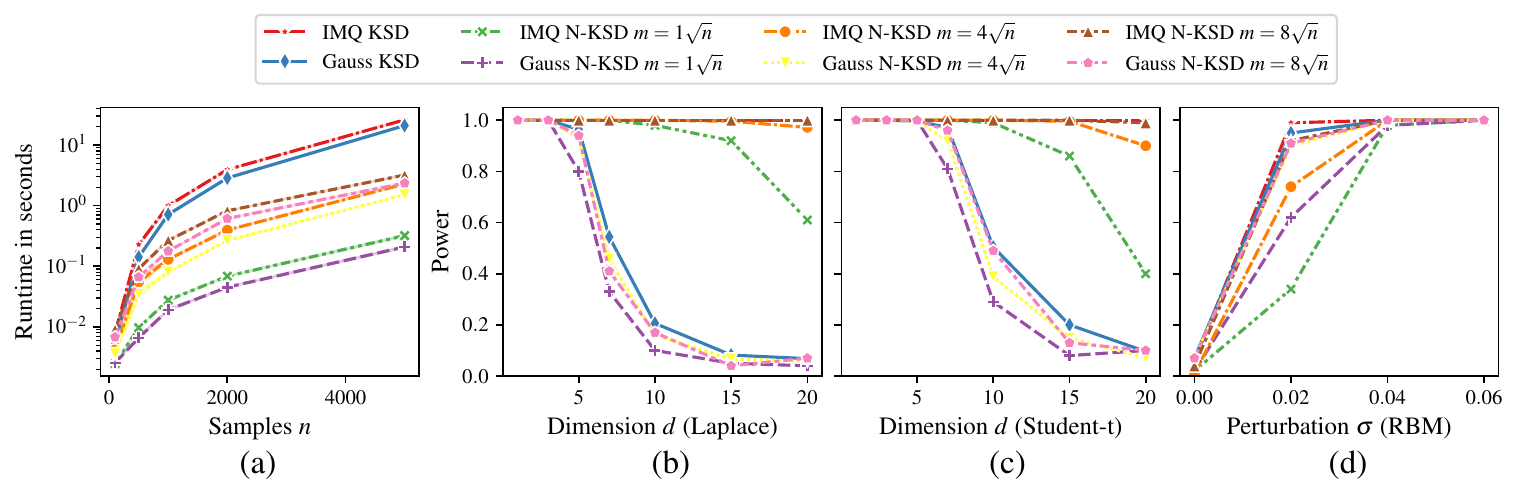}
    \caption{Impact of different choices of factor $c$ for the number of Nyström samples $m=c\sqrt n$. }
    \label{fig:impact-of-m}
\end{figure}

Figure~\ref{fig:impact-of-m}(a--d) captures the impact of the choice of Nyström samples $m=c\sqrt n$ for $c\in\{1,4,8\}$; the $\sqrt n$ dependence follows from Corollary~\ref{corr:decay-assumption}(ii), where we neglect the logarithmic terms due to their small contribution. We include the quadratic time approaches as baselines; the experimental setup matches the experiments detailed in the article in Section~\ref{sec:experiments}. Generally, as one expects, both runtime and power increase for larger $c$. Still, even for $c=8$, where the power of the proposed approximation is hardly discernible from the baselines across all experiments, its runtime is an order of magnitude lower, which further strengthens the benefit of employing our proposed method.

\vfill

\end{document}